\documentclass[iicol,sn-nature]{sn-jnl}

\usepackage{graphicx}
\usepackage{multirow}
\usepackage{amsmath,amssymb,amsfonts}
\usepackage{amsthm}
\usepackage{mathrsfs}
\usepackage[title]{appendix}
\usepackage{xcolor}
\usepackage{textcomp}
\usepackage{manyfoot}
\usepackage{booktabs}
\usepackage{algorithm}
\usepackage{algorithmicx}
\usepackage{algpseudocode}
\usepackage{listings}

\usepackage{subcaption}
\usepackage{mdframed}
\usepackage{svg}
\usepackage{xspace}

\usepackage{geometry}

\usepackage{lineno}

\theoremstyle{thmstyleone}

\theoremstyle{thmstyletwo}

\theoremstyle{thmstylethree}

\newcommand{\curia}{Curia\xspace}
\newcommand{\curiabench}{CuriaBench\xspace}

\newcommand{\reff}[2]{\hyperref[#1]{\ref*{#1}#2}.}

\begin{document}

\title[CURIA]{Curia: A Multi-Modal Foundation Model for Radiology}

\author*[1]{\fnm{Corentin} \sur{Dancette}}\email{corentin.dancette@raidium.eu}
\equalcont{These authors contributed equally to this work.}

\author[1,2,3]{\fnm{Julien} \sur{Khlaut}}
\equalcont{These authors contributed equally to this work.}

\author[1]{\fnm{Antoine} \sur{Saporta}}
\author[1,3,6]{\fnm{Helene} \sur{Philippe}}
\author[1]{\fnm{Elodie} \sur{Ferreres}}
\author[1]{\fnm{Baptiste} \sur{Callard}}
\author[1]{\fnm{Théo} \sur{Danielou}}
\author[1]{\fnm{Léo} \sur{Alberge}}
\author[1,6]{\fnm{Léo} \sur{Machado}}
\author[1]{\fnm{Daniel} \sur{Tordjman}}
\author[1]{\fnm{Julie} \sur{Dupuis}}
\author[1,2,3,4]{\fnm{Korentin} \sur{Le Floch}}
\author[7]{\fnm{Jean} \sur{Du Terrail}}
\author[8]{\fnm{Mariam} \sur{Moshiri}}
\author[9]{\fnm{Laurent} \sur{Dercle}}
\author[2,3,4]{\fnm{Tom} \sur{Boeken}}
\author[6]{\fnm{Jules} \sur{Gregory}}
\author[6]{\fnm{Maxime} \sur{Ronot}}
\author[10]{\fnm{François} \sur{Legou}}
\author[10]{\fnm{Pascal} \sur{Roux}}
\author[2,3,5]{\fnm{Marc} \sur{Sapoval}}
\author[1]{\fnm{Pierre} \sur{Manceron}}
\author[1]{\fnm{Paul} \sur{Hérent}}

\affil*[1]{\orgname{Raidium}, \orgaddress{\street{27 rue du faubourg Saint-Jacques}, \city{Paris}, \postcode{75014}, \country{France}}}
\affil[2]{\orgdiv{Department of Vascular and Oncological Interventional Radiology}, \orgname{Hôpital Européen Georges Pompidou, AP-HP}, \orgaddress{\city{Paris}, \country{France}}}
\affil[3]{\orgdiv{Faculté de Santé}, \orgname{Université Paris-Cité}, \orgaddress{\city{Paris}, \country{France}}}
\affil[4]{\orgdiv{HEKA}, \orgname{INRIA}, \orgaddress{\city{Paris}, \country{France}}}
\affil[5]{\orgdiv{PARCC U 970}, \orgname{INSERM}, \orgaddress{\city{Paris}, \country{France}}}
\affil[6]{\orgdiv{Department of Radiology, FHU MOSAIC}, \orgname{Beaujon Hospital}, \orgaddress{\street{APHP.Nord}, \city{Clichy}, \country{France}}}
\affil[7]{\orgname{.omics}, \orgaddress{\city{Paris}, \country{France}}}
\affil[8]{\orgdiv{Department of Radiology and Radiological Science}, \orgname{Medical University of South Carolina}, \orgaddress{\city{Charleston}, \state{SC}, \country{USA}}}
\affil[9]{\orgdiv{Department of Radiology}, \orgname{Columbia University Irving Medical Center}, \orgaddress{\city{New York}, \state{NY}, \postcode{10032}, \country{USA}}}
\affil[10]{\orgname{Centre Cardiologique du Nord}, \orgaddress{\city{Saint-Denis}, \postcode{93200}, \country{France}}}

\abstract{

AI-assisted radiological interpretation is based on predominantly narrow, single-task models. This approach is impractical for covering the vast spectrum of imaging modalities, diseases, and radiological findings. Foundation models (FMs) hold the promise of broad generalization across modalities and in low-data settings. However, this potential has remained largely unrealized in radiology.
We introduce Curia, a foundation model trained on the entire cross-sectional imaging output of a major hospital over several years—which to our knowledge is the largest such corpus of real-world data—encompassing 150,000 exams (130 TB).  On a newly curated 19-task external validation benchmark, Curia accurately identifies organs, detects conditions like brain hemorrhages and myocardial infarctions, and predicts outcomes in tumor staging. Curia meets or surpasses the performance of radiologists and recent foundation models, and exhibits clinically significant emergent properties in cross-modality, and low-data regimes. To accelerate progress, we release our base model's weights at~\url{https://huggingface.co/raidium/curia}.

}


\maketitle

\section{Introduction}\label{sec1}

\begin{figure*}[h]
    \centering
    \includegraphics[width=0.93\linewidth]{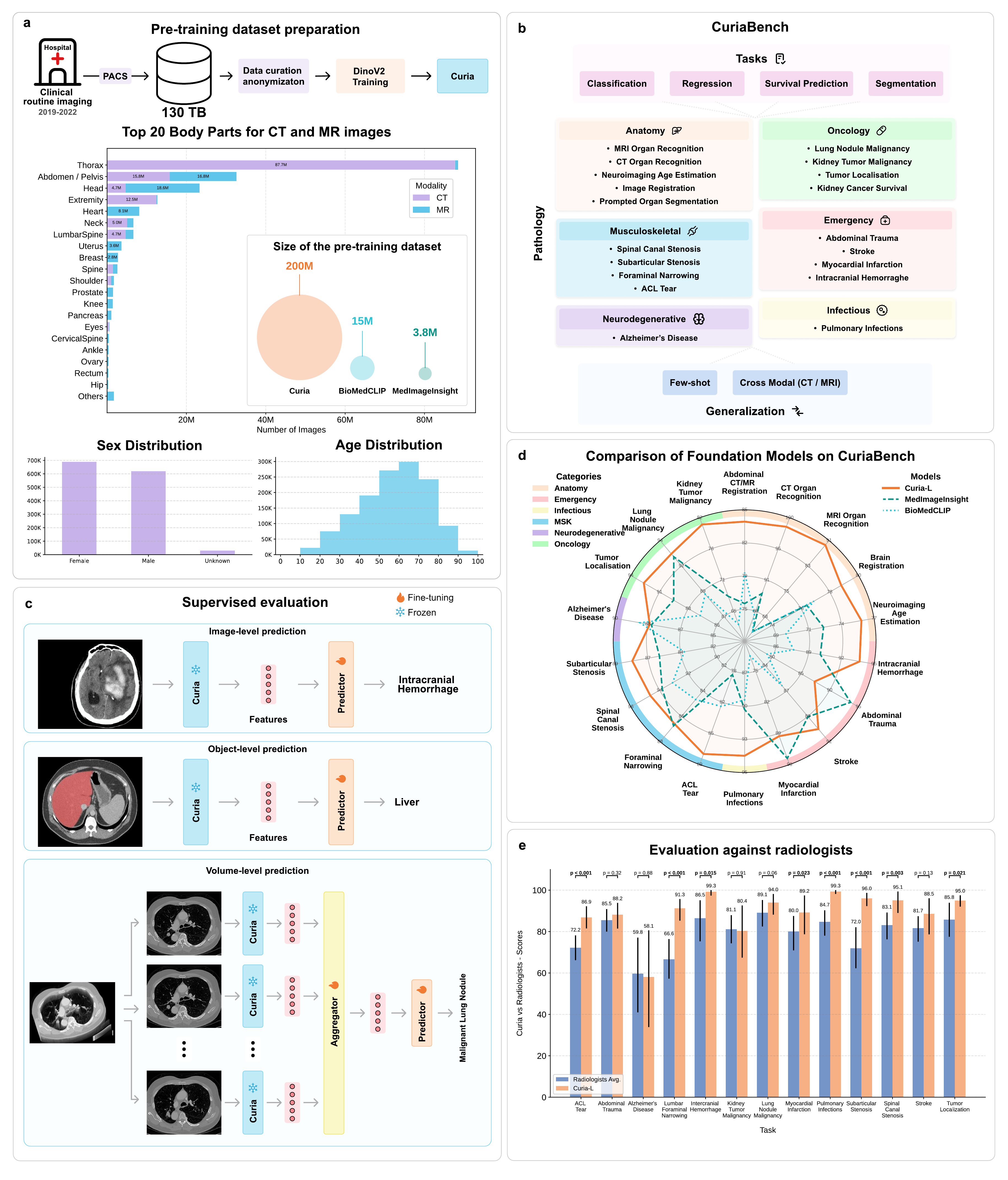}
    \caption{\small \textbf{Overview of \curia.} \curia is a radiological foundation model for CT and MRI images, trained with self-supervised learning with the DINOv2 algorithm on 200M images, based on the vision transformer architecture. \textbf{(a)} Pre-training methodology and statistics - All reported numbers correspond to the number of 2D images. PACS = Picture Archiving and Communication System. \textbf{(b)} List of tasks and pathology areas evaluated in the benchmark. We evaluated \curia on classification, regression, survival prediction, and segmentation tasks, and also explore generalization in few-shot and cross-modal settings. \textbf{(c)} Method for supervised evaluations: image-level prediction, object-level prediction, and volume-level prediction. \textbf{(d)} Radar plot of \curia-L's performance against MedImageInsight and BioMedCLIP.  Metrics are detailed in Fig.~\ref{fig:downstream-tasks}. To provide robust estimates, we report the mean performance over 1000 bootstrap samples for each task. \textbf{(e)} Performance comparison of \curia-L against resident radiologists. We report the mean performance with 95\% confidence intervals, calculated over 1000 bootstrap samples, along with the statistical significance using a paired bootstrap hypothesis test.}
    \label{fig:main-figure}
\end{figure*}

Radiology is at the center of many medical specialties, which rely on radiologists' interpretation of images from various modalities, including CT, MRI, ultrasound, and X-ray \cite{article}.
The analysis of these images is crucial for detecting and characterizing medical conditions, quantifying disease progression, and monitoring treatment efficacy across a broad spectrum of diseases.
AI has the potential to enhance radiology workflows and improve radiologists' efficiency, particularly for labor-intensive tasks such as image segmentation, or specialized and/or complex tasks which are prone to inter-reader variability \cite{DBLP:journals/corr/LitjensKBSCGLGS17, diagnostics15091146}.
To date, the dominant paradigm in radiological AI development has involved training specialized models for individual tasks such as segmentation, abnormality detection (e.g., tumor detection), or pathology classification.
However, this ``one-task, one-model'' approach is exceptionally resource-intensive, as it necessitates the curation and manual annotation of large, task-specific datasets for each modality and clinical application \cite{prevedello2019challenges, bian2025artificial}.
It is potentially one of the bottlenecks in moving AI radiology models into the clinical workflow.

Foundation models (FM) represent a significant paradigm shift in the field of AI. More specifically, in the domain of natural images, self-supervised models such as DINOv2 \cite{oquab2023dinov2} and MAE \cite{he2022mae} have demonstrated the effectiveness of this approach, often reaching performances of supervised models. Leveraging large-scale unlabeled datasets, these models learn fine-grained semantic features that can be effectively transferred to downstream tasks using simple, lightweight classifiers with minimal or no fine-tuning.

Adapting these methods to medical imaging is a promising solution to tackle the plethora of radiological use cases across multiple image modalities.
By assisting radiologists in detecting and characterizing diseases, these models may help improve patient outcomes and streamline clinical workflows. Ultimately, integrating FMs into radiology offers a path toward enhanced diagnostic precision, innovative research, and personalized precision medicine~\cite{paschali2025foundation, lesaunier2025artificial}.
Previous research on FMs in radiology includes models such as BiomedCLIP \cite{zhang2025BioMedCLIP}, BiomedParse \cite{zhao2024biomedparse}, and MedImageInsight \cite{codella2024MedImageInsight}. These models have been trained on medium-scale datasets (e.g., 15M images for BiomedCLIP), larger than typical supervised training datasets, but considerably smaller than those used for FMs in natural language or vision (e.g., 120M images for DINOv2).
More critically, their training sets are often heterogeneous mixtures of biomedical images from various medical specialties (including ophthalmologic imaging, pathology, radiology, endoscopy, and dermatology). Because these datasets typically aggregate specialized collections, they can introduce biases that constrain the model's generalizability to novel scenarios and do not encompass the broad range of tasks a radiologist performs in its daily activities.
Adding to this challenge, the absence of a unified benchmark has prevented rigorous comparison of these existing FMs \cite{mahmood2025benchmarking}.

In this article, we apply self-supervised learning to a large-scale dataset of routine clinical cross-sectional imaging. Specifically, we pre-train vision transformer models (ViT-B and ViT-L \cite{vit}) on more than 200 million CT and MRI images (130 TB of data from 150K exams, see Fig.~\reff{fig:main-figure}{a}) using the DINOv2~\cite{oquab2023dinov2} algorithm.

Moreover, to assess the general performance of radiological FMs, we introduce a comprehensive benchmark, \curiabench, comprising 19 distinct radiological tasks (Fig.~\reff{fig:main-figure}{b}) that span both CT and MRI modalities and cover most anatomical regions. This benchmark encompasses a broad spectrum of clinical cases that a radiologist encounters, including disorders related to aging (\textit{e.g.}, Alzheimer's disease, degenerative spine condition), emergencies (\textit{e.g.}, Anterior Cruciate Ligament (ACL) tears, abdominal trauma, brain hemorrhage), infectious diseases (\textit{e.g.}, lung infections), and oncological conditions with survival predictors (\textit{e.g.}, renal malignancy). It contains classification, regression, survival prediction and segmentation tasks, and allows us to explore generalization in few-shot and cross-modal settings. We evaluate \curia without any fine-tuning, and only train lightweight prediction heads with the model's features (Fig.~\reff{fig:main-figure}{c}).

Our model, which we have named \curia, sets a new standard in radiological image interpretation. We present the result of \curia-B and \curia-L based on the ViT-B (for Base) and ViT-L (for Large) architectures \cite{vit}. Evaluation on our benchmark shows that \curia is highly adaptable across numerous tasks, performs strongly in few-shot learning scenarios, and demonstrates emergent cross-modal generalization capabilities -- \curia learns similar features for the same structures across modalities. Furthermore, the model consistently and significantly outperforms existing foundation models, such as BiomedCLIP and MedImageInsight (Fig.~\reff{fig:main-figure}{d}).
Notably, \curia delivers performance comparable to, or even exceeding, the accuracy of resident radiologists on the benchmark tasks (Fig.~\reff{fig:main-figure}{e}).

Our analysis is first structured to assess the model's generalization capabilities on anatomical tasks, focusing on few-shot learning and cross-modality performance. We then evaluate its performance on our benchmark of medical tasks.
We also conduct an in-depth study in oncology, which demonstrates that the model can help predict risks associated to tumors and their related survival rates, and we study the attention maps of the prediction heads, giving interpretability to the model's predictions.
Those results highlight the potential of \curia to accelerate the development of robust, versatile, and data-efficient AI tools to enhance patient care, ultimately equipping the community with powerful, novel models that deliver tangible clinical impact.

\section{Results}
A key aspect of FMs is their ability to adapt to a multitude of downstream tasks with minimal task-specific fine-tuning.
After the initial pre-training of \curia using the DINOv2 framework, our primary evaluation protocol consisted of training a lightweight classifier on the features extracted from the frozen model backbone.
More details about the methodology can be found in the Method Section~\hyperref[sec:method]{\ref{sec:method}}.

We compared \curia against two other FMs:
\begin{itemize}
    \item MedImageInsight~\cite{codella_medimageinsight_2024}, an open-source visual embedding model by Microsoft, trained on multi-modal medical data from various domains (radiology, histology, pathology, dermatology, ophthalmology), for a total of 3.8M images. It is based on the DaViT~\cite{ding2022davit} architecture, and contains 360M parameters.
    \item BiomedCLIP~\cite{zhang_biomedclip_2025}, a ViT-B model trained with contrastive learning on 15M (image, text) pairs extracted from PubMed.
\end{itemize}

\begin{figure*}[!h]
    \centering
    \includegraphics[width=0.8\linewidth]{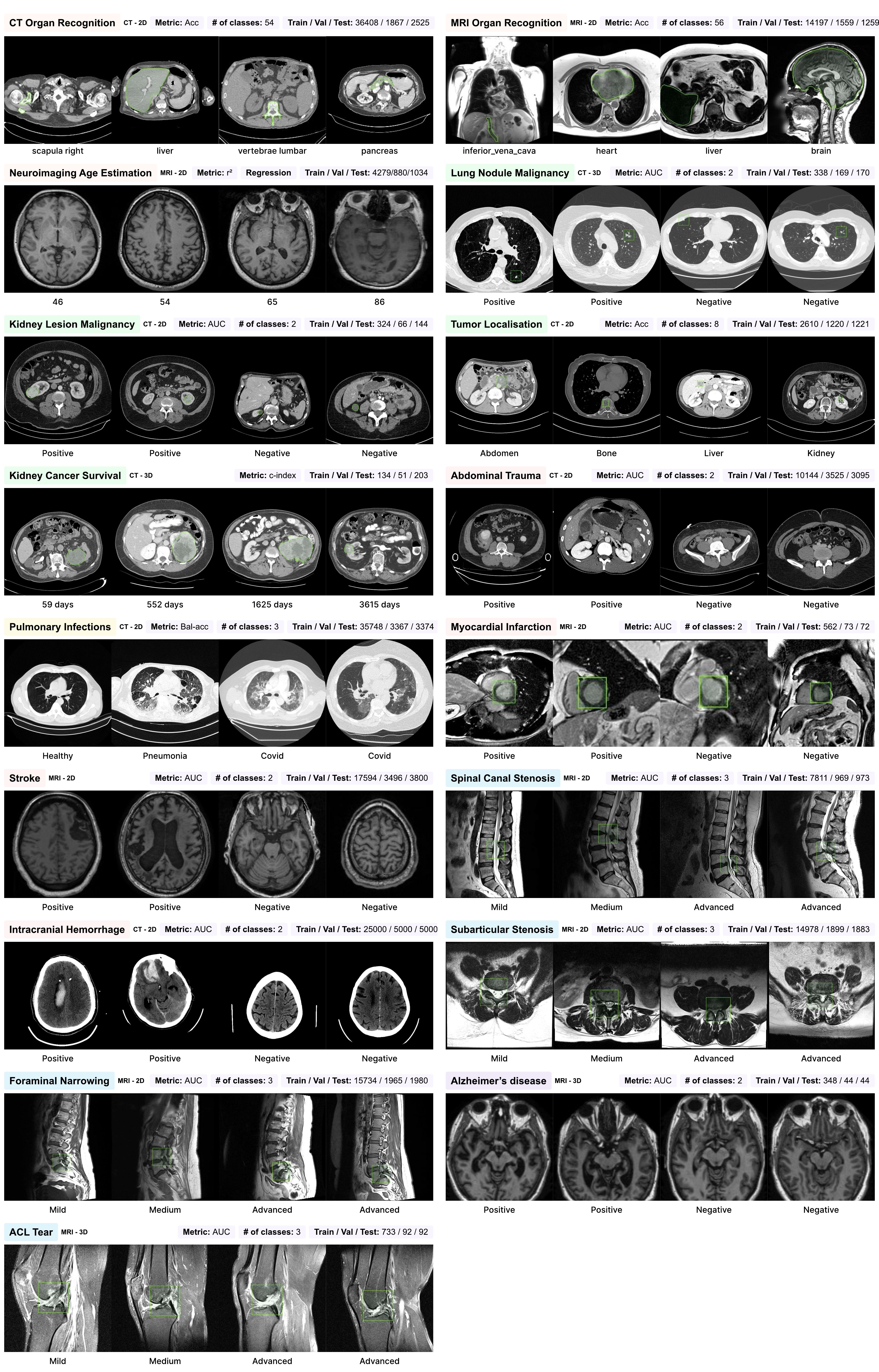}
    \caption{\textbf{List of downstream tasks considered in the \curiabench benchmark.} For each task, we report the modality (CT/MRI), the type (2D/3D), the metric (Accuracy, AUC, Balanced Accuracy, $r^2$), the number of classes for classification tasks, and the sizes of the training, validation, and test sets. The registration task and the prompted segmentation task are showcased in Fig.~\ref{fig:2}.}
    \label{fig:downstream-tasks}
\end{figure*}

We present a summary of the main results from our benchmark in Fig.~\hyperref[fig:main-figure]{\ref*{fig:main-figure}d}. More details about the benchmark can be found in the Benchmark Section~\hyperref[sec:benchmark]{\ref{sec:benchmark}}. We also present examples of the whole benchmark in Fig. \hyperref[fig:downstream-tasks]{\ref{fig:downstream-tasks}}.

\subsection{Evaluation on Anatomical Benchmark}

To comprehensively evaluate \curia, we first established its proficiency in \textbf{organ recognition} across various body regions. We then investigated its data efficiency by evaluating our model in a few-shot setting on these tasks. A key focus of our study was to assess Curia's \textbf{cross-modality generalization}. By training on CT scans and evaluating on MRI scans, we investigated whether the model could capture fundamental, modality-agnostic features. To further demonstrate its capabilities, we examined its performance on \textbf{registration}, a task that inherently requires a deep understanding of spatial anatomy. We finally evaluated Curia on \textbf{prompted organ segmentation}.

\begin{figure*}[t!]
    \centering
    \includegraphics[width=0.95\linewidth]{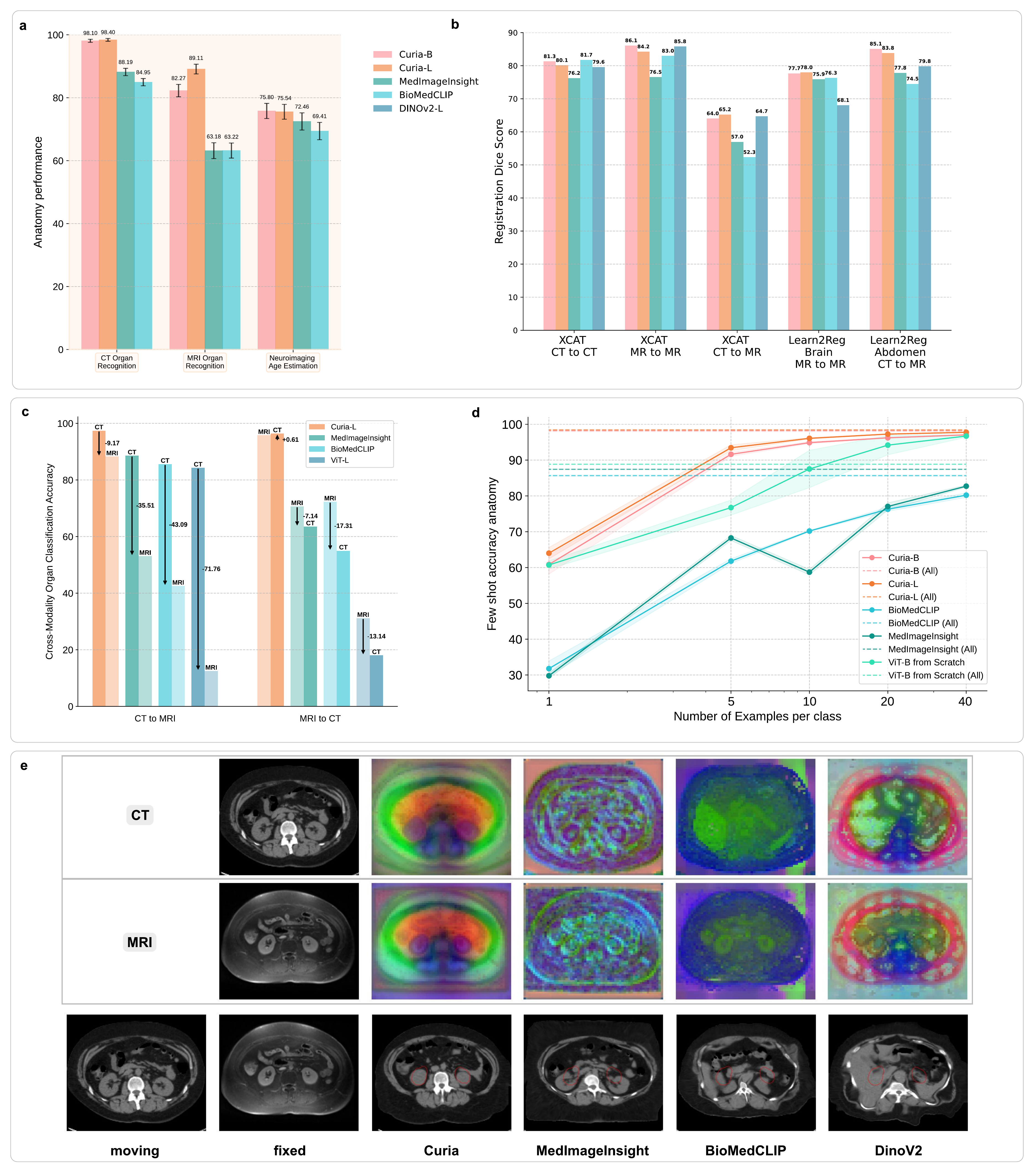}
    \caption{\small \textbf{Performance on anatomical tasks} \textbf{(a)}~Comparison of foundation models on the CuriaBench anatomical subset. Metrics are defined in Fig.~\ref{fig:downstream-tasks}. The error bars represent the 95\% confidence interval derived from 1000 bootstrap samples. \textbf{(b)} Performance on Imaging Registration on three datasets: XCAT, Learn2Reg Brain, Learn2Reg Abdomen. \textbf{(c)} Cross-modality generalization on organ classification. We report the gap between the two modalities for each model. \textbf{(d)} Data Efficiency: performance of FMs and model from scratch with varying number of labeled samples on the \textit{Anatomy - CT} task. ``All'' are models trained on the full dataset. \textbf{(e)} First two lines: Principal Component Analysis (PCA) visualization of feature maps from \curia, MedImageInsight, BiomedCLIP, and DINOv2 on a CT and an MRI image. Last line: Image registration results using image features. A displacement field was computed, and used to project the \textit{moving} image to match the \textit{fixed} image. We display the projected image for \curia, MedImageInsight, BioMedCLIP and DINOv2. We also show in red the positions of kidneys from the fixed image for reference.}
    \label{fig:2}
\end{figure*}

\paragraph{\curia obtains excellent performance in anatomy classification}

We evaluated models on organ classification in both CT and MRI images, based on our \textbf{CT Organ Recognition} and \textbf{MRI Organ Recognition} benchmarks.
\curia-L outperformed other FMs in organ classification on CT scans, achieving a near-perfect accuracy score of 98.40\% (Fig.~\hyperref[fig:2]{\ref*{fig:2}a}), outperforming both MedImageInsight, which achieved  88.19\% ($P<0.001$) and BiomedCLIP, which achieved 84.95\% ($P < 0.001$).
On MRI data, \curia-L obtained an accuracy of 89.11\% and also surpassed the other models: MedImageInsight with 63.18\% ($P < 0.001$) and BiomedCLIP with 63.22\% ($P < 0.001$).

\paragraph{\curia can predict the age from brain MRIs}
From a T1-weighted MRI image of a healthy patient's brain, the model was tasked with predicting the \textbf{patient's age}. Our model \curia-L achieved an age prediction RMSE of ±6.15 years with an $r^2$ score of 75.54 (Fig.~\hyperref[fig:2]{\ref*{fig:2}a}). In comparison, BiomedCLIP had an RMSE of ±7.18 years and an $r^2$ score of 69.41 ($P$ $<$ 0.001), whereas MedImageInsight obtained ±6.72 years with an $r^2$ score of 72.46 ($P$ = 0.004).

\paragraph{\curia is more efficient in the low-data regime than other FMs}
We investigated the \textbf{few-shot learning} capabilities of FMs using the CT Organ Recognition benchmark. Experiments were conducted with varying sample sizes per class, ranging from 1 to 40 images per class. In addition, we represent the performance of each model trained using all available data from the CT Organ Recognition benchmark with a dashed line. The results, presented in Fig.~\hyperref[fig:2]{\ref*{fig:2}d}, indicate that \curia's performance is near its maximum accuracy with a small number of training samples.
MedImageInsight exhibited lower performance compared to \curia across various sample sizes, with its final accuracy being 10.2 percentage points lower than \curia's, and the performance gap was more pronounced at 20 samples per class, resulting in an approximately 20-point difference between the two models. A model trained from scratch required 10 examples per class to reach more than 80\% accuracy.

\paragraph{\curia displays emerging properties of cross-modal generalization}

We evaluated \curia's ability to generalize across different imaging modalities by training a linear classification head for anatomy recognition on CT scans and evaluating it on MRI scans, and vice versa, based on the Cross-Modality Organ Recognition benchmark.
We also performed the same experiment for MedImageInsight, BiomedCLIP, and a ViT-L trained from scratch, and present the results in Fig.~\hyperref[fig:2]{\ref*{fig:2}c}.

On CT to MRI, \curia demonstrated a cross-modal generalization capability, exhibiting a balanced accuracy decrease of 9.17 percentage points when evaluated on the out-of-distribution MRI dataset. Notably, \curia achieved higher accuracy on MRI in this zero-shot setting than other foundation models when trained directly on MRI. In contrast, other  FMs showed larger drops in performance, ranging from 35.51 to 71.76 percentage points. As anticipated by the absence of pre-training, a ViT-L model trained from scratch suffered the most pronounced performance degradation, highlighting its limited ability to generalize to the target modality.

On MRI to CT, although other models exhibited more moderate performance drops--ranging from 7.14 to 17.31--the key finding was that \curia maintained virtually identical performance between the in-distribution MRI dataset and out-of-distribution CT dataset. Remarkably, its accuracy even improved slightly by 0.61 percentage points, underscoring its robust generalization ability across image modalities.

\paragraph{\curia allows better volume registration across modalities than other FMs}

The results in Fig.~\hyperref[fig:2]{\ref*{fig:2}b}, complemented by the per-organ Dice Similarity Coefficient (DSC) values in Table~\ref{tab:registration-gan-mrct} for XCAT and Table~\ref{tab:registration-learn2reg-mrct} for Learn2Reg Abdomen, demonstrate that \curia consistently outperformed or matched the performance of other models across all \textbf{image registration} tasks. For XCAT CT-to-CT registration, \curia-B and \curia-L achieved mean DSC of 81.30\% and 80.12\%, respectively, fairly close to BiomedCLIP's performance on the task with 81.74\%. All three models excelled in liver (94.26\%, 93.37\%, and 94.65\% for \curia-B, \curia-L, and BiomedCLIP, respectively) and spleen (87.18\%, 87.30\%, and 90.22\% for \curia-B, \curia-L, and BiomedCLIP, respectively). In XCAT MR-to-MR registration, \curia led with a mean DSC of 86.10\% and 84.25\% for \curia-B and \curia-L, respectively, achieving the best scores across all organs.

Similarly to organ classification, we explored cross-modal capabilities of \curia for image registration. For XCAT CT-to-MR registration, \curia achieved the highest mean DSC (64.03\% and 65.25\% for \curia-B and \curia-L, respectively) and outperformed others on all organ-specific metrics, with a notable liver DSC of 86.12\% for \curia-B and 85.34\% for \curia-L. Although MedImageInsight and BiomedCLIP performed well in certain areas, they were less consistent, particularly in cross-modality tasks.
On Learn2Reg benchmarks, \curia consistently outperformed other models. More specifically, on Learn2Reg Abdomen MRI/CT, \curia-B and \curia-L achieved mean DSC scores of 85.1\% and 83.84\%, respectively, greatly surpassing the performance of both MedImageInsight and BiomedCLIP, achieving 77.83\% and 74.52\%, respectively. On Learn2Reg Brain, the difference between model performance is less pronounced, but \curia-B and \curia-L still led with mean DSC scores of 77.68\% and 77.96\%, respectively, compared to MedImageInsight's 75.91\% and BiomedCLIP's 76.29\%.

Furthermore, we experimented with DINOv2 Large as a baseline for image registration. Despite not being trained on medical images, it achieved respectable performance across benchmarks, at times even outperforming MedImageInsight and BiomedCLIP or matching \curia's performance.

Finally, it is also worth noting that \curia maintained competitive results across benchmarks on smoothness metric measured by the standard deviation values of the log of the Jacobian determinant (stdLogJ)~\cite{hering2022learn2reg}.

To further investigate the behavior of the different FMs on image registration, we performed PCA visualizations of their extracted feature maps. Specifically, we projected the high-dimensional features onto a 2D space to qualitatively assess the semantic alignment between modalities. These visualizations were generated on an MRI image and its corresponding registered CT images. The results, shown in Fig.~\hyperref[fig:2]{\ref*{fig:2}e}, reveal distinct structural patterns in the feature embeddings, offering insights into how well each model captures anatomical consistency across modalities.
Interestingly, while MedImageInsight and BiomedCLIP delineated anatomical regions to some extent, their projections predominantly exhibited one or two dominant colors, indicating limited variation across principal components. This suggests a lower diversity in their feature representations. In contrast, Curia’s feature maps displayed a broader range of colors corresponding to different anatomical structures. This increased visual complexity reflects a richer and more discriminative embedding space, aligning with Curia’s stronger quantitative performance in registration tasks.

\begin{figure*}[t]
    \centering
    \includegraphics[width=0.97\linewidth]{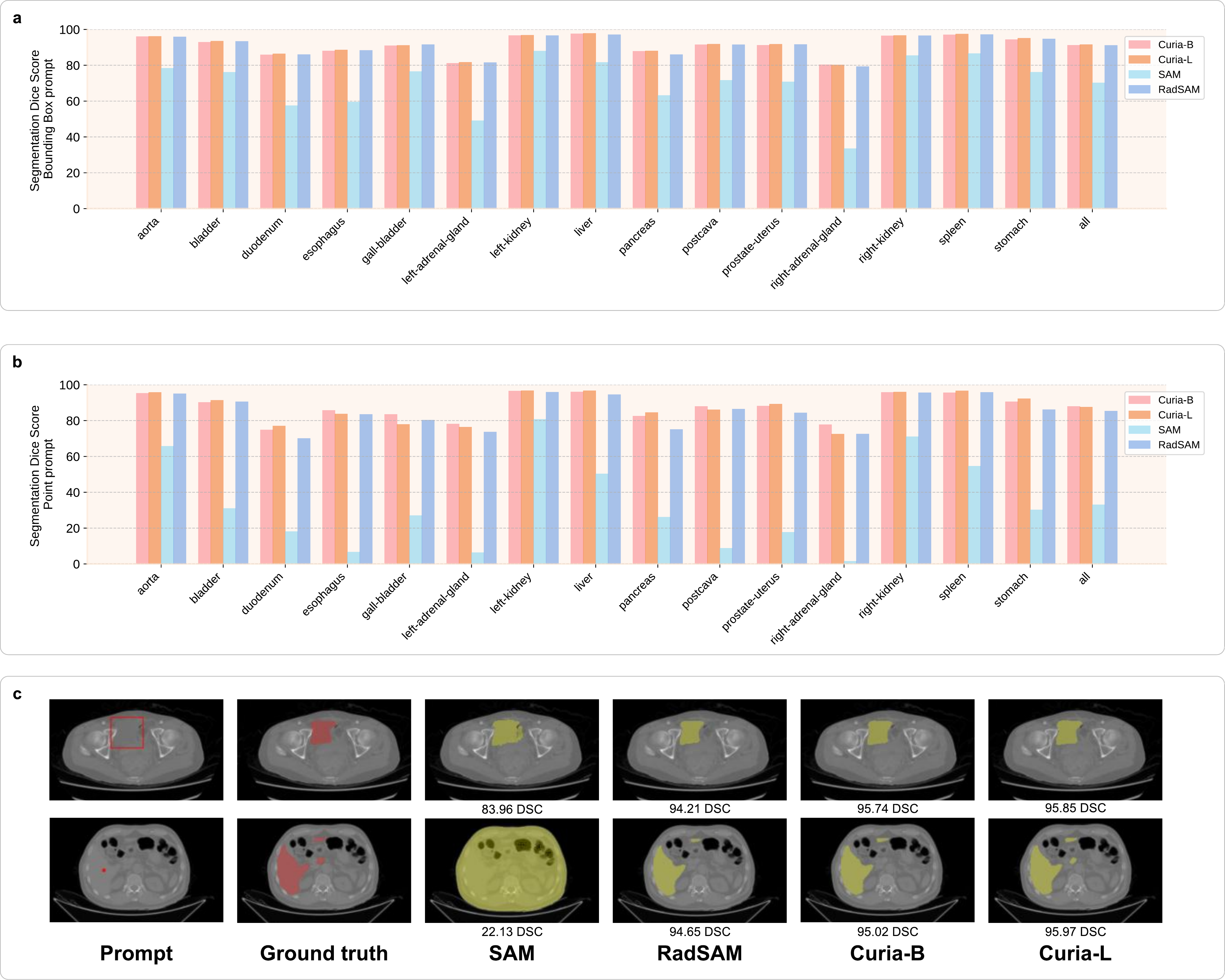}
    \caption{\textbf{Performance of Curia, SAM and RadSAM on the Prompted Segmentation benchmark.} \textbf{(a)}~Performance with a bounding box prompt. For each structure, we report the average Dice Score over all samples. \textbf{(b)}~Performance with a point prompt. For each structure, we report the average Dice Score over all samples.  \textbf{(c)}~Visualization of the segmentation maps predicted by \curia-B and \curia-L with the SAM decoder. We compare the results with SAM and RadSAM on the same image and prompt. Top row: with a bounding box prompt; Bottom row: with a point prompt.}
    \label{fig:prompted-segmentation}
\end{figure*}

\paragraph{\curia can be adapted for prompted segmentation, matching the performance of specialized models}

We compared \curia in the \textbf{prompted segmentation} framework~\cite{sam} for radiological images, similar to available models such as MedSAM~\cite{medsam} or RadSAM~\cite{radsam}. We conducted the same evaluation protocol as used in RadSAM, employing both point and bounding box prompts on the Prompted Organ Segmentation benchmark. We replaced the original SAM vision encoder with our \curia backbone and performed a two-stage fine-tuning process. We finally evaluated our approach against SAM and RadSAM (Fig.\reff{fig:prompted-segmentation}{a},\reff{fig:prompted-segmentation}{b}).

Our performance results were on par with RadSAM. Using bounding box and point prompts, the RadSAM model achieved DSC scores of 91.08\% and 85.27\%, respectively. In comparison, our \curia-L model obtained DSC scores of 91.49\% and 87.49\% while \curia-B got 91.13\% and 87.87\%. We also evaluated the original SAM model, which was pre-trained on approximately 1 billion masks from 11M non-medical images. It achieved significantly lower DSC scores of 70.16\% and 33.04\% with bounding box and point prompts, respectively, highlighting the importance of designing segmentation methods specifically for the medical domain. In Supplementary Table C3, we also report the DSC scores per organ. These results demonstrate the quality of \curia's features, which has not been pre-trained for segmentation but attained results comparable or better than RadSAM.

Qualitative results are presented in Fig.~\reff{fig:prompted-segmentation}{c}, showing the predictions of each model. \curia-L successfully segmented all disconnected components of the liver using only a single point prompt. RadSAM and \curia-B were accurate on the main regions but failed to capture one component. With a bounding box prompt, all models produced similar segmentation masks, except for the original SAM model, which again failed to generate accurate segmentation, as in the example with the point prompt.

\begin{figure*}[t]
    \centering
    \includegraphics[width=0.92\linewidth]{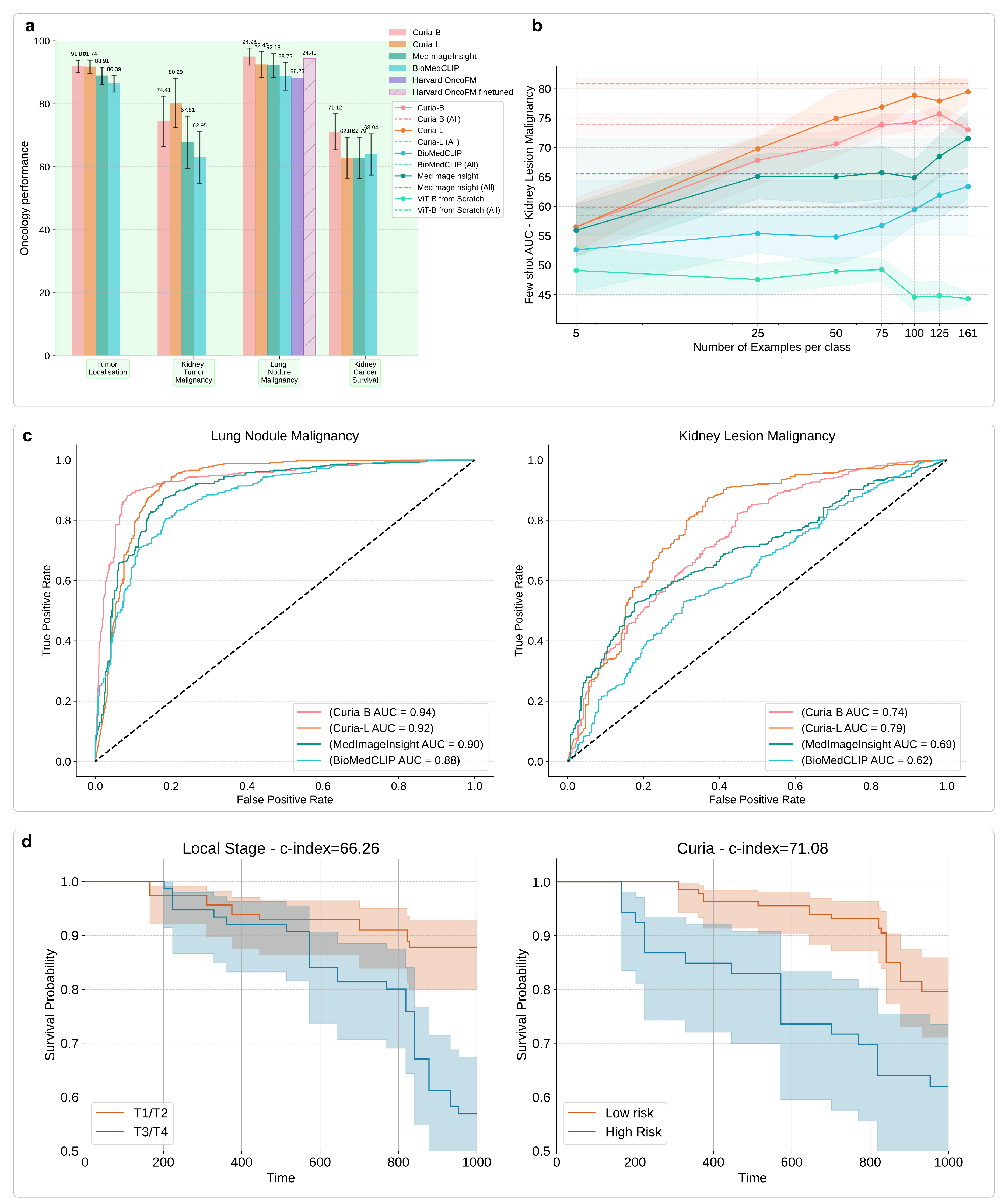}
    \caption{\small \textbf{Performance of \curia on Oncology-related tasks}
    \textbf{(a)} Results of CuriaBench Oncology subset -- Tumor anatomical site, kidney tumor and lung nodule malignancy, and renal malignancy survival. We compared \curia against BioMedCLIP and MedImageInsight, as well as Harvard Onco-FM~\cite{pai_foundation_2024} for lung nodule malignancy.
    The 95\% CI of the scores obtained with 1000 bootstrap samples is shown by error bars, except for Harvard-RT for which we report the score from the original publication.
    \textbf{(b)} Performance (AUC) of \curia, BioMedCLIP, MedImageInsight, and a ViT-B without any pre-training, with varying number of examples in the training set of the Kidney lesion malignancy task. The error bars represent the variance over 5 runs for each point.
    \textbf{(c)} ROC curves for Lung nodule and Kidney lesion malignancy tasks. All models were evaluated 5 times, and we aggregated the predictions using the pooling method~\cite{swets2012evaluation, hogan2023averaging} to display the final ROC curve.
    \textbf{(d)} Kaplan–Meier curves for groups, stratified by local stage (T1/T2 vs T3/T4), and by model's risk prediction. The error bars represent the 95\% CI of the estimates.}
    \label{fig:3}
\end{figure*}

\subsection{Evaluation on Oncology Benchmark}

\paragraph{\curia outperforms generalist and specialized FMs in oncological tasks}

To investigate \curia's performance in oncology, we evaluated the task of finding the \textbf{tumor's localisation} (Fig.~\hyperref[fig:3]{\ref*{fig:3}a}).
\curia-L achieved a balanced accuracy of 91.74\%, while BiomedCLIP and MedImageInsight attained a balanced accuracy of 86.39\% ($P$ $<$ 0.001 ) and 88.91\% ($P$ = 0.041), respectively.

We then evaluated our models on two tumor classification tasks for kidney lesion and lung nodules where the aim is to predict the malingnancy of the tumor. We report the aggregate scores in Fig.~\hyperref[fig:3]{\ref*{fig:3}a} and we also show the ROC curves in Fig.~\hyperref[fig:3]{\ref*{fig:3}c}.

On \textbf{kidney lesion malignancy} classification, \curia-B achieved an average AUROC of 74.41 and \curia-L 80.29. This result surpassed the score achieved by other foundational models, with BiomedCLIP attaining 62.95 (\curia-B $P$ $<$ 0.001, \curia-L $P$ $<$ 0.001) and MedImageInsight achieving 67.81 (\curia-B $P$ = 0.177, \curia-L $P$ = 0.025). We display the ROC curve in Extended Fig.~\ref{fig:roc}.

Regarding \textbf{lung nodule malignancy}, \curia-B obtained an average AUROC of 94.98, and \curia-L 92.45  while MedImageInsight and BiomedCLIP obtained comparable scores of 92.18 (\curia-B $P$ = 0.482, \curia-L $P$ = 0.993) and 88.72 (\curia-B $P$ = 0.126, \curia-L $P$ = 0.876), respectively. Additionally, we also compared to the original paper by Pai et al.~\cite{pai2024foundation}, using their dataset split. Their Onco-FM obtains an AUROC of 94.40 with full fine-tuning, and 88.23 with a feature-based approach. \curia outperformed these results, without any fine-tuning of the base model. We display the ROC curve in Extended Fig.~\ref{fig:roc}.

Finally, we investigated the low-data regime, as illustrated in Fig.~\hyperref[fig:3]{\ref*{fig:3}b}. Similar to the anatomical tasks, we evaluated the FMs on the Kidney lesion malignancy task using varying numbers of training examples, and compared their performance to FMs fine-tuned on the full dataset (dashed lines). While we see high variances with a low number of examples(5-25), \curia-B and \curia-L's performance increased greatly with additional examples, outperforming the other models by a large margin, when trained with more than 50 examples per class. The model trained from scratch did not learn to classify kidney lesions at all with a small number of examples, obtaining an AUC of around 0.5.

\paragraph{\curia helps predict survival rate in cancer patients}

To probe the capacity of our FM for complex clinical reasoning, we tackled the challenging problem of oncologic prognosis, focusing on the prediction of \textbf{survival time} at baseline, using a cox regression model~\cite{cox1972regression, monod2024torchsurv} on the model's features.
On a cohort of 183 patients with renal malignancies from TCIA~\cite{Clark2013tcia}, resident radiologists annotated the lesion positions with pixel-level masks.
We then trained the cox regression model~\cite{monod2024torchsurv} to predict the survival time utilizing the concordance index (c-index) as a readout.
We benchmarked the image‑based survival predictor against conventional tumor staging using the T-stage of the TNM classification, also known as the local stage. Tumors were categorized into low-stage (T1–T2) and high-stage (T3–T4) groups according to their locoregional spread.

\curia-B and \curia-L achieved a c-index of 0.71 and 0.63, respectively, for survival prediction. Notably, \curia-B substantially outperformed both BiomedCLIP (c-index: 0.64, \curia-B $P$ = 0.035, \curia-L $P$ = 0.79) and MedImageInsight (c-index: 0.63, \curia-B $P$ = 0.003, \curia-L $P$ = 0.99) on the benchmark. The local stage alone obtained a c-index of 0.66. The plot of the two Kaplan-Meier curves is shown in Fig.~\hyperref[fig:3]{\ref*{fig:3}d}.

\begin{figure*}
    \centering
    \includegraphics[width=1.0\linewidth]{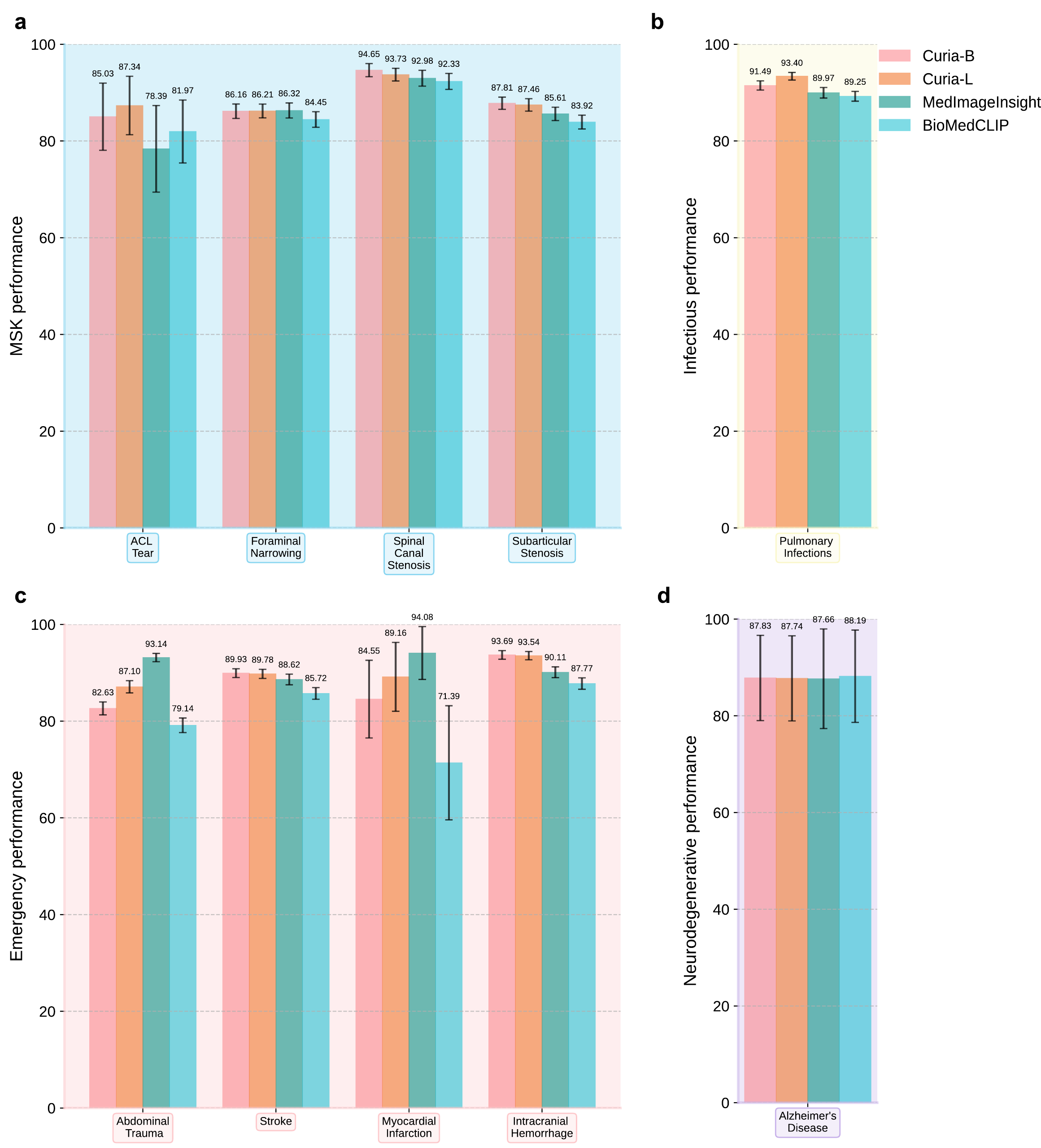}
    \caption{\textbf{Performance of \curia, BioMedCLIP and MedImageInsight on four subsets of CuriaBench.} Metrics are described in Fig.~\ref{fig:downstream-tasks}. The metrics reported are the average across 1000 bootstrap samples for 5 runs, and we show the 95\% confidence intervals.
    \textbf{(a)} Musculoskeletal (MSK) conditions
    \textbf{(b)} Infectious conditions
    \textbf{(c)} Emergency-related conditions
    \textbf{(d)} Neurodegenerative conditions
    }
    \label{fig:4}
\end{figure*}

\subsection{Evaluation on Musculoskeletal Benchmark}

\paragraph{\curia achieves leading performance in musculoskeletal disease assessment}
The model accurately classified the severity of \textbf{degenerative disease of the lumbar spine}. It was able to assess three types of conditions:  foraminal narrowing, subarticular stenosis, and spinal canal stenosis, defined through three severity levels (Mild, Moderate, or Severe). \curia-L obtained AUROC scores of 86.21 for foraminal narrowing, 87.46 for subarticular stenosis, and 93.73 for spinal canal stenosis. As shown in Fig.~\hyperref[fig:4]{\ref*{fig:4}a}. Compared to other FMs, \curia-L obtained comparable or better performance on foraminal narrowing -- 84.45 for BiomedCLIP ($P$ = 0.07) and 86.32 for MedImageInsight ($P$ = 0.87) -- and equivalent performance on spinal cord stenosis -- 92.33 for BiomedCLIP ($P$ = 0.344) and 92.98 for MedImageInsight ($P$ = 0.243). Notably, \curia-L outperformed both models on subarticular stenosis: BiomedCLIP achieved 83.92 ($P$ = 0.02) and MedImageInsight achieved 85.61 ($P$ $<$ 0.001).

We also studied the performance of \curia on \textbf{ACL tear} benchmark in knee MRI. By showing a cropped region of interest around the ligament of interest, \curia-L obtained an AUROC of 87.34, whereas BiomedCLIP and MedImageInsight achieved significantly lower scores with 81.97 ($P$ = 0.004) and 78.39 ($P$ = 0.013), respectively (Fig.~\hyperref[fig:4]{\ref*{fig:4}a}).

\subsection{Evaluation on Emergency Benchmark}

\paragraph{\curia delivers competitive results in emergency medicine }
Fig.~\hyperref[fig:4]{\ref*{fig:4}c} shows the performance of \curia on multiple emergency-related medical tasks.
First, the model was able to detect the presence or absence of \textbf{intracranial hemorrhage on head CT examinations}. \curia-L reached an AUROC of 93.54 on the test set. In comparison, MedImageInsight and BiomedCLIP achieved lower AUROC scores of 90.11 ($P$ $<$ 0.001) and 87.77 ($P$ = 0.015), respectively.

\curia accurately detected \textbf{myocardial infarction} in 2D cardiac MRI images. From a squared mask around the myocardium, \curia-L obtained an AUROC of 89.16. MedImageInsight obtained a higher score of 94.08 ($P$ = 0.104), while BiomedCLIP was significantly lower at 71.39 ($P$ $<$ 0.001).
\curia was also able to detect signs of \textbf{active intra-abdominal bleeding} on abdominal CT images.
\curia-L achieved an AUROC of 87.10, while MedImageInsight and BiomedCLIP obtained AUROCs of 93.14 ($P$ $<$ 0.001) and 79.14 ($P$ $<$ 0.001), respectively.
Finally, \curia could detect signs of past \textbf{strokes} on brain T1-weighted MR images. \curia-L obtained an AUROC of 89.78, while MedImageInsight and BiomedCLIP obtained 88.62 ($P$ = 0.001) and 85.72 ($P$ $<$ 0.001), respectively. We display the ROC curves in Extended Fig.~\ref{fig:roc} for those three tasks.

\subsection{Evaluation on Neurodegenerative Benchmark}

\paragraph{\curia is competitive on neurodegenerative disease}
We evaluated \curia on \textbf{Alzheimer's disease} prediction on brain MRI images.
On the full MRI volume, \curia-B and \curia-L obtained an AUROC of 87.83 and 87.74, respectively, whereas BiomedCLIP and MedImageInsight obtained scores of 88.19 (\curia-B $P$  = 0.003, \curia-L $P$  = 0.005) and 87.66 (\curia-B $P$ = 0.936, \curia-L $P$  = 0.325), respectively, as shown in Fig.~\hyperref[fig:4]{\ref*{fig:4}d}.

\subsection{Evaluation on Infectious Benchmark}

\paragraph{\curia achieves superior accuracy in pulmonary infection detection compared to previous FMs}

We evaluated \curia on \textbf{pulmonary infections} with our dedicated benchmark, which contained images of patients diagnosed as COVID-19 positive, non COVID pneumonia positive, or negative.
\curia-L obtained a balanced accuracy of 93.40\%, outperforming BiomedCLIP that obtained 89.25\% ($P$ $<$ 0.001) and MedImageInsight with 89.97\% ($P$ $<$ 0.001) as shown in Fig.~\hyperref[fig:4]{\ref*{fig:4}b}.

\subsection{Comparison to Radiologists}
\paragraph{\curia outperforms radiology residents on most tasks}

Fig.~\hyperref[fig:main-figure]{\ref*{fig:main-figure}e} compares the performance of \curia against the average scores of four final-year radiology residents across fourteen different medical imaging tasks.
More details on the radiologist evaluation method is given in the Methods Section~\ref{ssec:radiologist-eval}. The results show that \curia obtained performance comparable to, and often higher than, the radiologists' predictions.
Overall, the data indicate that \curia was reliable across a wide range of medical imaging tasks, suggesting that it could be a valuable tool to support clinical diagnosis and enhance diagnostic consistency.

\subsection{Interpretability of Curia's predictions}

\begin{figure*}
    \centering
    \includegraphics[width=1.0\linewidth]{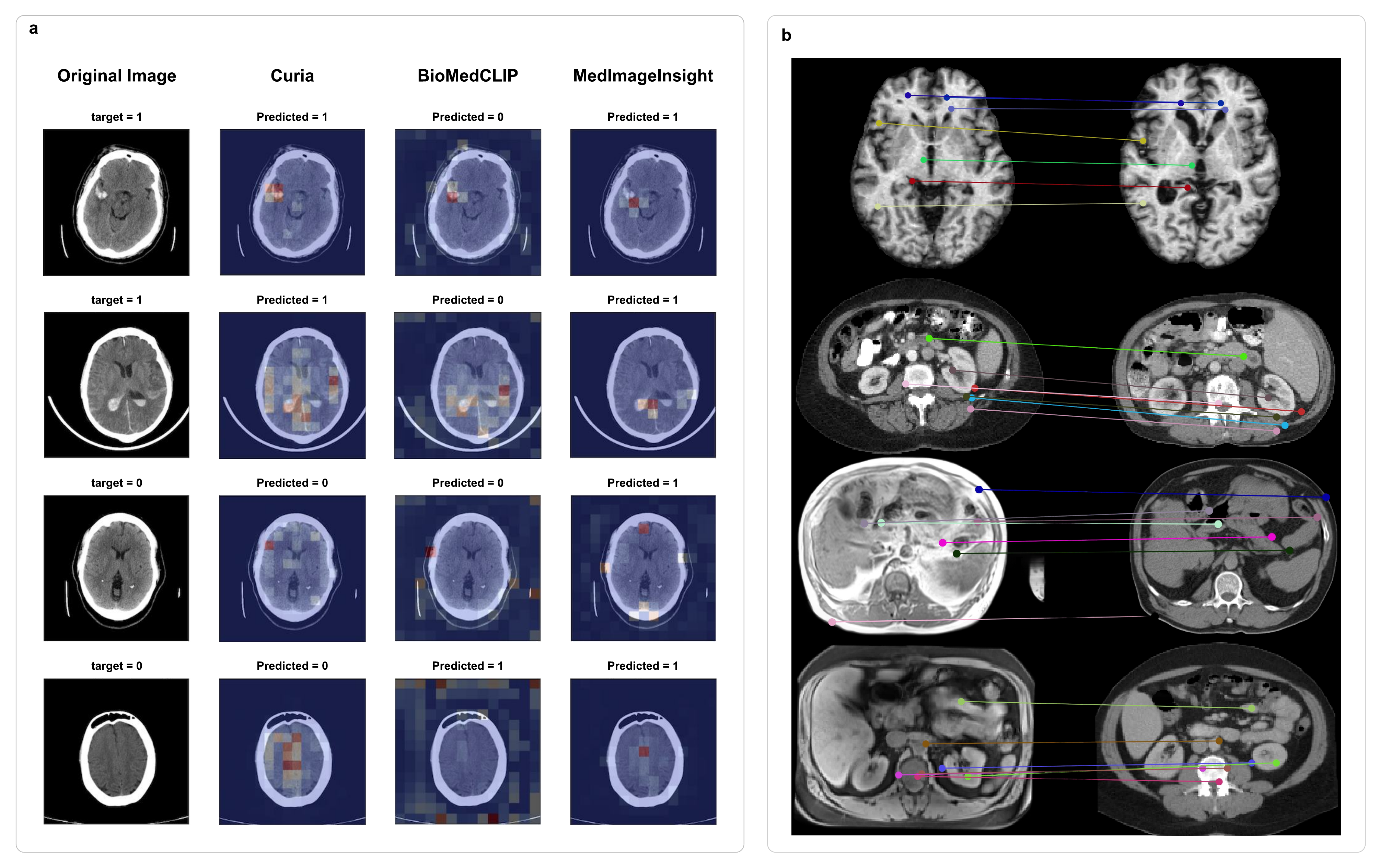}
    \caption{\textbf{(a)} Visualization of the attention maps. Images displayed are windowed to brain standard viewing parameters (level=40, width=80) with a varying blue-to-red colormap corresponding to increasing attention scores. The attention maps were computed between the final patches of each model and a single learnable query vector, highlighting the areas used in the decision-making process.
    \textbf{(b)} Visualization of keypoint matching. The first row illustrates keypoint matching between two MRI images from different patients in the OASIS dataset. The second row presents matches between two CT images from different patients in the Learn2Reg CT-Abdomen dataset. The third and fourth rows demonstrate cross-modality matching between MRI (source) and CT (target) images from the same patient in the Learn2Reg MR-CT dataset.}
    \label{fig:attention-maps-viz}
\end{figure*}

To qualitatively asses the focus and interpretability of the feature maps of FMs, we acquired the attention maps for \curia, BiomedCLIP, and MedImageInsight on the intracranial hemorrhage classification task. These maps, shown in Fig.~\hyperref[fig:attention-maps-viz]{\ref*{fig:attention-maps-viz}a}, represent the cross-attention weights of the best-performing classifiers, each trained with a single query vector. Negative instances yielded more diffuse attention patterns, aligning with the premise that no singular region is indicative of a negative finding. It was also observed that BiomedCLIP, which demonstrated the lowest classification accuracy, generated the most widespread attention maps, often encompassing areas beyond the anatomical boundaries of the brain.

To further acknowledge the robustness and generalization of \curia's feature representations, we performed patch-level keypoint matching using \curia-B between the features of a source and a target 2D image as shown in Fig.~\hyperref[fig:attention-maps-viz]{\ref*{fig:attention-maps-viz}b}. Keypoints are randomly sampled from the source image and matched to the most similar patches in the target image based on cosine similarity scores. We used three datasets from the image registration benchmark: OASIS~\cite{marcus2007open}, Learn2Reg CT-Abdomen and Learn2Reg-MR-CT~\cite{hering2022learn2reg}. We conducted this experiment under different setups: using MRI as the source and CT as the target from the same patient, as well as using source and target images from different patients but within the same modality. The results demonstrate \curia-B's ability to perform cross-modality and inter-patient feature transfer, highlighting its capacity to understand relationships between anatomically similar regions across different imaging modalities.

\subsection{Scaling curves}
Extended Fig.~\ref{fig:scaling} presents a series of experiments with varying dataset sizes and training durations for a subset of our benchmark tasks on the ViT-B and ViT-L architectures. These results highlight that both dataset size and training duration are important factors for downstream performance.

\section{Discussion}\label{sec12}

In this article, we introduced \curia, a multi-modality FM for radiology, with a comprehensive benchmark of 19 tasks to evaluate its capabilities. Our results demonstrate that by pre-training on a large-scale dataset of over 200 million unlabeled CT and MRI images, Curia established a new standard in radiological image interpretation, consistently outperforming existing models.

\vspace{10pt}
A major contribution of this study is the demonstration that self-supervised applied to a large, unlabeled dataset can produce a model with robust generalization capabilities. Unlike previous models trained on smaller, more specialized, and often heterogeneous biomedical datasets, \curia's training on a large body of routine clinical images has resulted in a deep, transferable understanding of complex anatomy and pathology. This is evidenced by its superior performance on a wide array of tasks spanning different anatomical regions (abdomen, brain, chest) and medical specialties, including oncology, musculoskeletal conditions, and emergency imaging.

\vspace{10pt}
One of the most significant findings is \curia's emergent property of cross-modal generalization.
The model, despite being trained on CT and MRI data without explicit pairing, can generalize features from one modality to another.
For instance, when trained for organ recognition on CT images, it demonstrates a strong ability to perform the same tasks on MRI, significantly outperforming other models, which suffer from substantial performance drops.
This suggests that Curia has learned modality-agnostic representations of anatomical structures, a critical step toward creating truly universal radiological AI.
This capability is further highlighted in registration tasks, where Curia excels at CT-to-CT, MR-to-MR, and even the more challenging cross-modality CT-to-MR alignments, maintaining high accuracy and plausible deformations.
Our experiments also show the data efficiency of the FM paradigm. Curia displays strong few-shot learning performance, achieving high accuracy on anatomical classification tasks with a small number of labeled examples. This is a crucial advantage in the medical imaging domain, where large, expertly annotated datasets are notoriously difficult and costly to produce.

\vspace{10pt}
In oncological imaging, we demonstrate that the model can automatically and efficiently characterize lesions, labeling them as benign or malignant, which could aid physicians. Additionally, the model exhibits strong performance on risk assessment, surpassing the score used in clinical practice to predict survival for kidney cancer, paving the way for complex FM-derived predictive biomarkers in oncology.

\vspace{10pt}
Although our findings are encouraging, this study has certain limitations.
First, while the pre-training dataset was large and diverse in content, the data source is from a single center, which may introduce institutional biases affecting generalizability, such as site‑specific imaging protocols, reliance on a specific vendor for the imaging equipment, or specificities of the local patient population.
However, the impact of this limitation is somewhat mitigated through utilization of a multi-center evaluation benchmark, which could support the claim of generalizability of \curia.
Second, \curia is fundamentally a 2D model, processing volumetric CT and MRI data on a image-by-image basis. While this approach is computationally efficient, it necessitates the addition of specialized prediction heads to aggregate 2D features for 3D tasks such as volumetric segmentation or registration. A native 3D FM could potentially offer improved performance on tasks that require a deeper characterization of volumetric images.
Lastly, while our benchmark includes 19 well-defined radiological tasks across CT and MRI, it does not yet cover the entire spectrum of imaging practice which also include ultrasound,  X-ray imaging, and nuclear medicine imaging, some of which are cornerstones of global diagnostic workflow. Expanding the benchmark to include additional imaging modalities and clinical tasks will be essential to further assess and extend the universality of models like \curia.
Finally, translating this significant technical achievement into a practical clinical asset is a complex endeavor where AI excellence is not the only requirement. Integration into hospital IT systems (PACS), adherence to strict regulatory standards, and acceptance within established physician workflows are other hurdles that need to be addressed to truly translate FM research into clinical use.

\vspace{10pt}

In conclusion, \curia represents a significant advancement in the application of FMs to radiology. By leveraging large-scale, self-supervised pre-training, it achieves leading performance across a diverse set of clinical tasks, demonstrates impressive data efficiency, and exhibits powerful cross-modal generalization. This study provides a robust foundation and a standardized benchmark for future research in the field, paving the way for the development of more powerful, versatile, and data-efficient AI tools that can enhance diagnostic accuracy and assist clinical workflows.
Looking forward, the evolution of Curia will likely center on incorporating rich, multimodal data from electronic health records and textual reports. Such an approach promises to unlock a deeper level of contextual understanding, significantly boosting generalization and enabling conversational interactions where clinicians can interact with the model using natural language.

\clearpage

\section{Methods}
\label{sec:method}
\subsection{Pre-training recipe}\label{sec:Pre-training recipe}
\paragraph{Pre-training dataset curation}
\label{sec:method:pretraining-dataset}

We partnered with a private hospital to create a dataset from routine cross sectional clinical examinations from 2019 to 2022. All exams were completely anonymized (all identifying metadata was removed, and defacing was applied on exams encompassing the patient's head). The original dataset contains 130TB of data, totaling 228M DICOM files (164M CT and 64M MR DICOM files).
To ensure high-quality data, only 3D CT and MR exams with at least 5 images were kept, and low-quality localizer or scout sequences were removed.
For our study on scaling curves, we created sub-versions of our dataset of different sizes: 30K, 200K, 2M, 20M and 200M images.

\paragraph{Large-Scale Pre-training}

\begin{figure*}[h]
    \centering
    \includegraphics[width=1\linewidth]{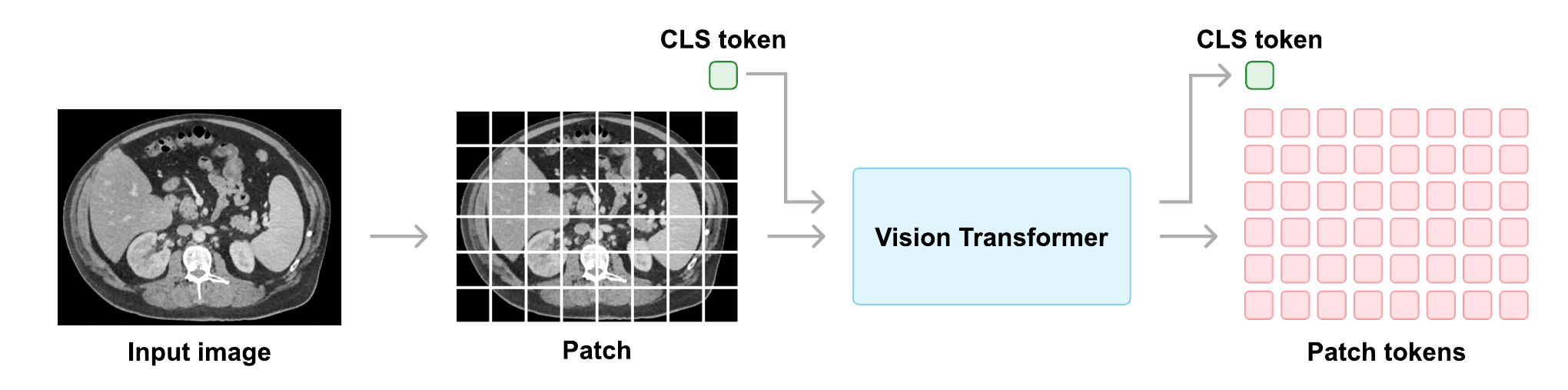}
    \caption{The Vision Transformer architecture. The image is tokenized into 16x16 patches, then converted into tokens and fed into the vision transformer. An additional class token is added, to perform image-level pre-training tasks.}
    \label{fig:vit}
\end{figure*}

\textbf{Preprocessing -- } All images were resized using bilinear interpolation to a fixed 512x512 dimension, and then normalized using z-score standardization. Input images are divided into 16x16 patches like shown in Fig. \hyperref[fig:vit]{\ref{fig:vit}}.
For CT images, we sampled all images in the axial axis.
For MRI, we sampled images following the acquisition axis.
For BiomedCLIP and MedImageInsight, we followed the pre-processing, and automatically applied windowing adapted to the task when possible (\curia was not trained with windowing, all images were processed with the same normalization). \\
\textbf{Architecture and training --} We used standard Vision Transformer~\cite{vit} models for the architecture of \curia. We adapted the DINOv2 codebase~\cite{oquab2023dinov2} for medical imaging. We trained two variants of this model: ViT-B, resulting in \curia-B, which contains 86M parameters, and ViT-L, resulting in \curia-L, which contains 300M parameters.
We trained the ViT using the self-supervised learning objective from DINOv2~\cite{oquab2023dinov2}. It is a combination of multiple losses: an image-level objective (aligning representations of class tokens between a teacher and a student network), a patch-level objective (based on masking random patches), and multiple regularization losses. We used DINOv2 default augmentations, except for the image rotations and color jittering, following \cite{moutakanni2024you} advocating for using only cropping for self-supervised learning. We trained our model on smaller datasets to find optimal learning rates and hyperparameters (learning rates and transforms). We report the final parameters in Table~B1.
Our final models were trained on 475,000 steps on a distributed cluster of 16 A100 GPUs for the ViT-B, and 32 A100 for the ViT-L.
Curia-B is trained on 20M images, while Curia-L on the full dataset of around 200M images.
The training time was approximately 5 days for the largest model.

\subsection{Evaluation setting}\label{sec:Evaluation setting}

\subsubsection{Adapting the model for downstream tasks}

\begin{figure*}[h]
    \centering
    \begin{subfigure}{1.0\textwidth}
    \includegraphics[width=1.0\linewidth]{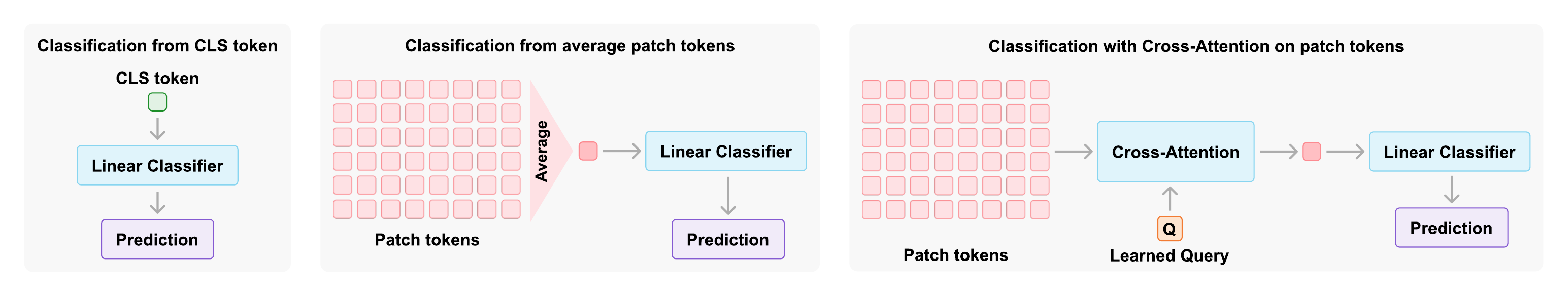}
    \caption{}
    \label{fig:classification-image}
    \end{subfigure}
    \begin{subfigure}{1.0\textwidth}
    \centering
    \includegraphics[width=1.0\linewidth]{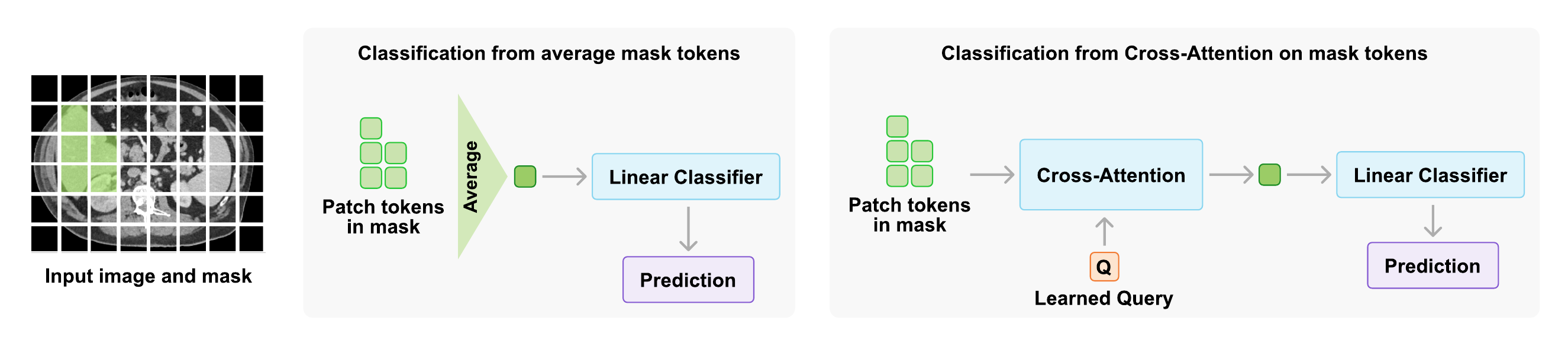}
    \caption{}
    \label{fig:classification-mask}
    \end{subfigure}
    \caption{Classification methods from the output of the vision transformer. \textbf{(a)} Image-level classification methods. We either used the class token, or pooled the patch tokens using an average or a cross-attention with a learned query. \textbf{(b)} Mask-level classification methods: we pooled the mask tokens, using either an average or a cross-attention with a learned query, and then performed linear classification.}
    \label{fig:classification-method}
\end{figure*}

To adapt the model for downstream tasks, we trained classification, regression, and survival heads on top of our FM, without fine-tuning the ViT weights. This allowed us to have a lightweight and fast adaptation for downstream tasks.

All heads, unless otherwise specified, were trained using stochastic gradient descent and a cosine scheduler, with a grid search of 10 learning rates. The best head was chosen on a held-out validation set and evaluated on the test set to obtain the final results.

For each task, we report the chosen head in Supplementary Table~B2.

\paragraph{Image-level classification tasks}

We evaluated multiple types of heads for classification tasks shown in Fig.~\hyperref[fig:classification-method]{\ref*{fig:classification-method}}.

\begin{enumerate}
    \item Classification from the class token: Similarly to DINOv2~\cite{oquab2023dinov2}, we trained a linear layer on top of the CLS token of the ViT. This is suitable for image-level classification tasks, but may fail if the task requires identifying fine-grained details in the image.
    \item Classification from patch tokens: we pooled all the class tokens together using an average or a max pooling, then trained a linear layer, similarly to the previous method.
    \item Attention-based pooling: we added a single cross-attention layer with a learned query to aggregate the patch tokens. The model could then learn to use specific parts of the image if necessary.
\end{enumerate}

\paragraph{Mask-level classification tasks}

Some tasks involve classifying a zone in the image -- for example, identifying a specific organ. Instead of cropping the image around the organ and feeding this crop to the vision transformer, we input the whole image. We then apply similar methods to the image-level classification tasks: apply an average pooling of all mask tokens and perform a linear classification, or use a cross-attention block followed by a linear classifier. We show in Fig.~\ref{fig:classification-mask} the two methods.

\paragraph{Handling 3D volumes tasks with a 2D image model}
For downstream tasks with 3D volumes, (e.g. lung nodule malignancy, or ACL tear), we forwarded all the images of the volume separately as detailed in Fig.~\reff{fig:main-figure}{c} We were then able to apply similar methods to the 2D case: in the volume-level classification (without mask) setup, we either pooled the patch tokens or the class tokens, and performed linear classification, or trained a cross-attention layer to perform the pooling.
In the mask-level classification setup, similarly, we performed an average pooling on the mask, or trained an attention head on the mask patches.

\paragraph{Adapting the model for regression downstream tasks}

We trained a regression head on top of class of mask tokens to perform regression tasks. We trained linear heads and multi-layer perception (MLP) heads, both with the mean squared error (MSE) loss. The MLP head was made of three layers, and the ReLU activation function and batch normalization were used between each layer.

\paragraph{Adapting the model for survival prediction}

To perform survival prediction, we added a linear head on top of the models and used the torchsurv package~\cite{monod2024torchsurv} to compute the negative partial log likelihood loss from the Cox survival framework. This loss took into account the censoring that happens in longitudinal data. The main metric used to perform model evaluation and selection was the c-index. Additionally, to separate the test samples into two groups, we chose a threshold that maximized the log-rank test statistic on a held-out validation set.

\paragraph{Adapting the model for registration tasks}

Zero-shot registration can be achieved using \curia or, equivalently, any model that can encode both global image information (class tokens) and patches (patch tokens). We followed the methodology described in the DINO-Reg paper ~\cite{song2024dino}. In their approach, class tokens are first used at the image level to perform rigid registration by estimating pairwise differences between images. Subsequently, deformable registration is achieved by using patch tokens as features in an optimization problem. This two-step adaptation enables the evaluation of the quality of both global (class tokens) and local (patch tokens) features. In order to obtain fine-grained deformation fields for \curia, and BiomedCLIP models, images were upsampled to 1024x1024 and 896x896, respectively, leading to 64x64 and 56x56 deformation fields, respectively. For MedImageInsight, this led to poor results (probably due to its different vision encoder architecture), and therefore, we chose the best configuration, which was to keep the models input dimensions but extract features after the first block leading to a 60x60 deformation field.

\paragraph{Adapting the model for prompted segmentation}

We followed the promptable segmentation paradigm established by recent works like SAM~\cite{sam} and RadSAM~\cite{radsam}. The architecture comprises three core components: (1) a vision encoder, (2) a prompt encoder, and (3) a mask decoder. For the vision encoder, we replaced the original SAM encoder with our \curia, which was initialized with its pre-trained weights. The prompt encoder and mask decoder architectures were adopted and initialized directly from SAM.
We employed a two-stage training procedure to effectively integrate the components. In the first stage, we froze the weights of the \curia vision encoder and trained only the prompt encoder and mask decoder. This allowed the model to learn the prompt-decoding mechanism without altering the powerful base features of \curia. In the second stage, we unfroze the vision encoder and performed an end-to-end fine-tuning of the entire model, allowing all parameters to adapt to the target segmentation task. In the first stage, we trained \curia-B and \curia-L for 6 epochs with a global batch size of 384 and a learning rate of $10^{-3}$. In the second stage, we trained \curia-B for 8 epochs and \curia-L for 6 epochs with a global batch size of 384 and a learning rate of $7.5e^{-5}$. All trainings were performed on 4 nodes with 4 80GB NVIDIA A100 GPUs.

\subsection{Statistical Analysis}

For all supervised training experiments, we performed non-parametric bootstrapping with 1,000 samples to report 95\% confidence intervals. Specifically, to better evaluate each model performance across training runs, we applied bootstrapping to the mean performance over 5 runs.
We report bootstrapping metrics on each benchmark in Appendix~\ref{app:scores}.
For statistical significance, we performed a two-sided paired bootstrap test with 1000 samples on the mean performance across 5 training runs for each model pair, in order to estimate the p-value. We report statistical significance results in Supplementary Table~\ref{app:pvalues}.

\subsection{Radiologist Evaluation}
\label{ssec:radiologist-eval}

We created a tool to evaluate radiologists on the benchmark tasks. For each task, the radiologists had the possibility to visualize training set examples along with their labels, change the windowing, and scroll through the images if the exam is in 3D. The training set label distribution was displayed.
They were then asked to annotate a subset of the testing set, on which a score was calculated.
When comparing with \curia in Fig.~\reff{fig:main-figure}{e}, we computed the metrics on the same subset for each task.
Our cohort of evaluators consisted in four resident radiologists in Paris-area hospitals.

\subsection{The \curiabench Benchmark}
\label{sec:benchmark}

This section describes \curiabench, a benchmark consisting in 19 downstream tasks we used to evaluate \curia and other FMs. Fig.~\ref{fig:downstream-tasks} presents example images for each task, detailing their modality, the number of images in the training, validation, and test sets, and the performance metric used in the benchmark.

\subsubsection{Anatomical Benchmark}

\paragraph{CT Organ Recognition}
\label{par:ct-rec-bench}
To create \textbf{CT Organ Recognition}, we used the Total Segmentator (TS)~\cite{totalsegmentator} dataset to create a benchmark for organ classification on CT scans. The task consisted of predicting the organ class based on the image and a mask of the organ.
We merged some classes from TS together, such as individual ribs and vertebrae, due to the difficulty of distinguishing them in 2D images.
We sampled one image from TS for each 3D volume and annotated organ pair, weighted by the number of mask pixels on each image.
The final dataset contained 54 organ labels. We used a part of the training set to build a held-out testing set. The training, validation and test sets contained 23,096, 1200 and 1554 samples, each containing one image-mask-target triplet.

\paragraph{MRI Organ Recognition}
\label{par:mri-rec-bench}
We used the same process with Total Segmentator MRI~\cite{d2024totalsegmentatormri} to create the \textbf{MRI Organ Recognition} task. It contained 56 classes.
The final dataset contains 14,197 training samples, 1559 validation samples and 1259 test samples.

\paragraph{Cross-Modality Organ Recognition}
\label{par:xm-rec-bench}
 For \textbf{Cross-Modality Organ Recognition} experiments, we follow a similar procedure to construct both CT and MRI benchmarks, but restrict the label space to the 41 anatomical classes shared across both modalities. This ensures consistent evaluation of generalization performance between modalities.
The final datasets contained 54,394 training samples, 2,718 validation samples and 4,996 test samples for CT, and 13,470 training samples, 1,412 validation and 1,412 test samples for MRI.

\paragraph{Neuroimaging Age Estimation}
\label{par:brain-age-bench}
\textbf{Neuroimaging Age Estimation} is a prediction task formulated as a regression problem using the IXI dataset \cite{ixidataset}. The dataset comprised approximately 600 MR images collected from normal, healthy individuals, with a mean age of 48 years (±16). For this task, we partitioned the dataset into 393 volumes for training, 80 for validation, and 94 for testing. Subsequently, we extracted 20\% of the brain's axial images from the T1-weighted MR images and trained the model on these images to predict the patient's age.

\paragraph{Image Registration}
\label{par:reg-bench}
Registration tasks evaluate the models' fine-grained representations at the patch level. FMs should ensure patch-level feature consistency across patients (anatomical registration), time (longitudinal tracking), and modalities (cross-modality alignment) enabling respectively population analysis/generalization, disease progression tracking, and multimodal support.
Three registration tasks were used to evaluate the proposed model and comparison to existing models: two tasks from the Learn2Reg challenge~\cite{hering2022learn2reg} and one synthetic multi-modal task. These tasks were evaluated in a zero-shot manner, meaning that no additional training or fine-tuning of the model was performed.

\ \\\noindent \textbf{Learn2Reg Abdomen MRI/CT - Intra-Patient Registration. }
\label{par:reg-abd-mri-ct-bench}
The Learn2Reg Abdomen MRI/CT task included 8 pairs of corresponding MRI and CT images from the same patients, sourced from the TCIA database~\cite{Clark2013tcia}. Data were resampled to an isotropic resolution of 2 mm, with dimensions standardized to 192×160×192.
Ground truth segmentations of the liver, spleen, and left and right kidneys were provided to evaluate registration performance. The evaluation was based on two metrics: the Dice Similarity Coefficient (DSC) to assess overlap accuracy and the standard deviation of the logarithm of the Jacobian determinant (SDlogJ), which evaluates the smoothness and plausibility of the displacement field.

\ \\\noindent \textbf{Learn2Reg Brain - Inter-Patient Registration.}
\label{par:reg-oasis-bench}
The Learn2Reg Brain task focused on whole-brain MRI registration, using data from the Open Access Series of Imaging Studies (OASIS). A total of 20 pairs of inter-patient T1-weighted MRI scans were selected to evaluate the model’s capability to align brain structures across subjects.
Anatomical segmentation labels for 35 brain structures were provided for evaluation. The data were preprocessed, including skull stripping, and resampled to an isotropic resolution of 1 mm with dimensions standardized to 160×192×224. This task emphasizes the model's ability to capture fine-grained structural information within the brain. Registration performance was measured using the DSC and SDlogJ metrics.

\ \\\noindent \textbf{XCAT - Synthetic Multimodal Abdominal Image Registration.}
\label{par:reg-synth-bench}
This task used a dataset generated synthetically by \cite{bauer2021generation} based on XCAT phantom data. A CycleGAN model was trained to map between the XCAT phantom and real image domains, producing synthetic T1-weighted MRI and CT images. The dataset comprised 56 inter-patient image pairs in both inhaled and exhaled states.
The data were pre-processed to an isotropic resolution of 2 mm and standardized to dimensions of 192×160×192. The exhaled phase served as the fixed reference, with the task requiring the registration of inhaled to exhaled phases. For cross-modality evaluation, the MRI images were used as the fixed reference.
Ground truth segmentations for the liver, spleen, and kidneys were generated using the TotalSegmentator tool \cite{totalsegmentator} on both T1-weighted and CT images. The evaluation relied on DSC to quantify overlap and SDlogJ to assess deformation plausibility.

\paragraph{Prompted Organ Segmentation}
\label{par:seg-bench}

For \textbf{Prompted Organ Segmentation}, we constructed our benchmark following the evaluation protocol established by RadSAM~\cite{radsam}. We utilized a subset of the AMOS dataset~\cite{amos}, which contained only CT images, for both training and evaluation. The final dataset comprised 300 CT scans with pixel-level annotations, covering 15  abdominal organs: spleen, kidneys (left and right), adrenal glands (left and right), gallbladder, esophagus, liver, stomach, aorta, postcava, pancreas, bladder, duodenum, and prostate/uterus. Evaluation was conducted by synthetically generating prompts in the form of bounding boxes and points for each 2D image. The bounding boxes were derived by perturbing the ground-truth boxes, introducing random offsets ranging from –5 to +20 pixels on each side. For the points, a random location was selected from within the ground truth mask, avoiding 2 pixels along its contour. The idea was to imitate prompts made by an operator.

\subsubsection{Oncology Benchmark}

\paragraph{Lung Nodule Malignancy}
\label{par:lung-tumor-bench}
We employed the LUNA16 dataset~\cite{setio2017luna16} with the specific split proposed by Harvard Onco-FM~\cite{pai2024foundation} to build the \textbf{Lung Nodule Malignancy} benchmark. This dataset comprised images of benign or suspicious pulmonary nodules. For our binary classification task, we utilized a 3D Region Of Interest (ROI) around each lesion. Notably, all ROIs had the same size, which simplified the analysis. The resulting datasets had 338 training samples, 169 validation samples, and 170 test samples.

\paragraph{Kidney Lesion Malignancy}
\label{par:kidney-tumor-bench}
We used the KITS23 dataset~\cite{heller2023kits21} to create the \textbf{Kidney Lesion Malignancy} benchmark. The objective of the task was classifying kidney lesions with two classes: solid tumors and cysts. We kept the mask annotations for the downstream task.
We randomly sampled one image per sample where the tumor or cyst mask was not empty. The sampling method ensured an even distribution across mask sizes. Finally, to balance the dataset, we ensured there were as many images for cysts and tumors. The resulting dataset had 324 training samples, 66 validation samples, and 144 test samples.

\paragraph{Tumor Localisation}
\label{par:tumor-site-bench}
We used the DeepLesion dataset~\cite{yan2018deeplesion} to establish \textbf{Tumor Localisation}, a benchmark for classifying the anatomical location of tumors. Specifically, given a 2D CT image and a region of interest (ROI) surrounding a tumor, the model was tasked with predicting the anatomical region type of the tumor (e.g., classifying if the tumor was located in the abdomen, bone, kidney, liver, lung, mediastinum, pelvis, or soft tissue). The ROIs corresponded to the bounding boxes provided in the original dataset. For this benchmark, one image per lesion was sampled from the dataset, excluding instances where the anatomical region was unspecified. The resulting dataset comprised 2,610 training samples, 1,220 validation samples, and 1,221 test samples, all with corresponding ROIs.

\paragraph{Kidney Cancer Survival}
\label{par:survival-bench}
To build the \textbf{Kidney Cancer Survival} benchmark, we assembled a kidney‑cancer cohort of 183 patients by extracting the molecular–clinical records of TCGA~\cite{tcga2008} from the imaging collections of TCIA~\cite{Clark2013tcia}.
The joint resource provided contrast‑enhanced CT scans, TNM staging and overall survival.
Data were retrieved with \texttt{tcia\_utils} Python client\footnote{\href{https://github.com/kirbyju/TCIA_Notebooks/blob/main/TCGA/TCGA_Clinical.ipynb}{TCIA\_Notebooks \texttt{TCGA\_Clinical.ipynb}}}  and converted from DICOM to NIfTI via dcm2niix \footnote{\url{https://github.com/rordenlab/dcm2niix}}.
We retained only those cases that carried either \texttt{days\_to\_death} or \texttt{days\_to\_last\_follow\_up} metadata.
Tumor volumes were semi‑automatically delineated with a segmentation FM~\cite{machado2025oncopilot}; two radiology residents ($>$ 2 years of oncology experience) then manually corrected the masks. The polished masks were fed to the FM whose classification head was replaced by a Cox layer to predict time‑to‑event, following our survival framework implementation.
We benchmarked the image‑based survival predictor against conventional anatomical staging using the T stage of the TNM classification, called local stage.

\subsubsection{Musculoskeletal Benchmark}

\paragraph{Degenerative Lumbar Spine}
\label{par:lumbar-spine-bench}
The Degenerative Lumbar Spine classification benchmarks were based on the dataset from the RSNA 2024 Lumbar Spine Degenerative Classification Challenge~\cite{rsna-2024-lumbar-spine-degenerative-classification}. This dataset consisted of distinguishing between five lumbar spine degenerative conditions which occur at intervertebral disc levels and are visible on specific MRI sequences:
\begin{itemize}
    \item Left and Right Foraminal Space Narrowing, visible on sagittal T1WI.
    \item Left and Right Subarticular Stenosis, visible on axial T2WI.
    \item Spinal Canal Stenosis, visible on sagittal T2WI and STIR.
\end{itemize}
The dataset provided severity scores -- that could take the values Normal, Moderate, or Severe -- for each imaging study in the dataset and each combination of medical condition and inter-vertebral disc level. The location of the anatomical sites where the conditions could occur were also available for every patient regardless of the presence of a medical condition, given as the coordinate of a 3D point on the corresponding MRI sequence. We constructed three benchmarks from this dataset.

\ \\\noindent \textbf{Foraminal Narrowing.}\label{par:nfn-bench} For each sagittal T1WI sequence, we selected the images on the sagittal axis based on the location of the anatomical sites given in the dataset. The objective of the benchmark was to predict the severity of foraminal narrowing with one of three values -- Normal, Moderate, or Severe. The benchmark dataset consisted in 31,468 training samples, 3930 validation samples, and 3960 test samples.

\ \\\noindent \textbf{Subarticular Stenosis.}\label{par:ss-bench} For each axial T2WI sequence, we selected the images on the axial axis based on the location of the anatomical sites given in the dataset. The objective of the benchmark was to predict the severity of subarticular stenosis with one of three values -- Normal, Moderate, or Severe. The benchmark dataset consisted in 29,956 training samples, 3798 validation samples, and 3766 test samples.

\ \\\noindent \textbf{Spinal Canal Stenosis.}\label{par:scs-bench} For each sagittal T2WI and STIR sequence, we selected the images on the sagittal axis based on the location of the anatomical sites given in the dataset. The objective of the benchmark was to predict the severity of spinal canal stenosis with one of three values -- Normal, Moderate, or Severe. The benchmark dataset consisted in 15,622 training samples, 1938 validation samples, and 1946 test samples.

\paragraph{Anterior Cruciate Ligament (ACL) Tear}
\label{par:acl-tear-bench}
The \textbf{ACL Tear} task \cite{vstajduhar2017semi} involved classifying MRI images of both knees to detect the presence or absence of an Anterior Cruciate Ligament (ACL) tear. Each knee volume was classified as one of the following: absence, injury, or complete rupture of the ACL.
To prepare this dataset for training, we selected a 3D box region within each volume where the ACL should be visible. We then evaluated the performance of models on the injury and complete Rupture classes using the ROC Area Under the Curve (AUC) metric.
The resulting datasets had 733 training samples, 92 validation samples, and 92 test samples. Notably, the sets were imbalanced; for instance, the training set had a significantly larger number of absences (554 samples) compared to injury (133 samples) and complete rupture (48 samples).

\subsubsection{Emergency Benchmark}

\paragraph{Myocardial Infarction}
\label{par:infarction-bench}
\textbf{Myocardial Infarction} was a classification task consisting in, given a cardiac MRI and a square mask around the myocardium, detecting if signs of infarction are visible or not. The EMIDEC dataset~\cite{emidec} provided 3D segmentations of the myocardium, cardiac cavity, infarction and no-reflow regions. The square mask input was computed from the myocardium segmentation.

\paragraph{Abdominal Trauma}
\label{par:abdominal-trauma-bench}
The \textbf{Abdominal Trauma} task consisted of predicting the presence of active contrast extravasation on axial CT images. The dataset from the RSNA 2023 Abdominal Trauma Detection Challenge~\cite{rsna-2023-abdominal-trauma-detection} provided the information of every image index that showed active extravasation. For the sampling of the 2D axial images, every image with active extravasation was added to the dataset. To ensure the balance of the dataset, an equal amount of images were randomly chosen from the remaining images without active contrast extravasation.

\paragraph{Intracranial Hemorrhage}
\label{par:brain-hemorrhage-bench}
The \textbf{Intracranial Hemorrhage} task consisted of predicting whether hemorrhage was present in a given cranial 2D CT image regardless of its type (e.g., subdural, epidural, intraparenchymal). An equal number of positive and negative images (25 000 in total) were sampled for the training set from the original dataset \cite{flanders2020construction}. The validation and test sets (5000 images in each set) were sampled randomly from the original dataset without balancing.

\paragraph{Stroke}
\label{par:stroke-bench}
The task was originally proposed by the ATLAS R2.0 dataset (Anatomical Tracings of Lesions After Stroke) \cite{atlasdataset} involving segmentation of brain lesions in patients who have experienced a stroke. For \textbf{Stroke}, we simplified this task into a classification problem, to determine whether an axial image contained brain lesions resulting from a stroke. The dataset included 655 T1-weighted MRI exams, which we split into 459 exams for training, 98 exams for validation, and 98 exams for testing. We extracted the axial images containing the brain and used 30\% of these images to create the dataset.

\subsubsection{Neurodegenerative Benchmark}

\paragraph{Alzheimer's Disease}
\label{par:alzheimer-bench}
This task was based on the Oasis-1 dataset \cite{marcus2007open}, which contained brain MRIs from patients with varying levels of dementia. The Clinical Dementia Rating (CDR) scale categorizes patients as: non-demented, very mild dementia, mild dementia, or moderate dementia.
For \textbf{Alzheimer's Disease} benchmark, we simplified the classification problem to a binary decision: either non-dementia or one of the other three dementia categories. We used the entire brain MRI volume for the classification pipeline rather than extracting specific features or regions of interest.
The resulting datasets were imbalanced, with 348 training samples, 44 validation samples, and 44 test samples. The significant disparity across classes is worth noting.

\subsubsection{Infectious Benchmark}

\paragraph{Pulmonary Infections}
\label{par:covid-bench}
For the \textbf{Pulmonary Infections} benchmark, we utilized the COVIDx CT dataset \cite{Gunraj2022}, which comprises chest CT scans of patients diagnosed as COVID-19 positive, non-COVID pneumonia positive, or negative (healthy). We created a stratified, sub-sampled version of the COVIDx CT dataset, maintaining the same training and evaluation splits. The resulting dataset represented 10\% of the original, consisting of 35,748 training samples, 3,367 validation samples, and 3,374 test samples. The training dataset was imbalanced, with significantly more COVID-19 positive cases compared to the other classes. However, this imbalance was mitigated in the validation and test sets which were more balanced.

\subsection{Computing software and hardware}

We used python for all experiments, with the Pytorch library, and the DINOv2 codebase~\cite{oquab2023dinov2} that we adapted for medical images.
We leveraged public HPC clusters to pre-train our model. For the ViT-B architecture, we used 4 nodes with 4 80GB NVIDIA A100 GPUs for 125 hours. We used DistributedDataParallel to train models with multi-GPU multi-node setting. All downstream experiments were done on a single NVIDIA 4090 GPU.
We used HuggingFace to load other FMs: BiomedCLIP (\url{https://huggingface.co/microsoft/BiomedCLIP-PubMedBERT_256-vit_base_patch16_224}, and MedImageInsight (\url{https://huggingface.co/lion-ai/MedImageInsights}).

\backmatter

\bmhead{Acknowledgements}

We acknowledge the EuroHPC Joint Undertaking for awarding this project access to the EuroHPC supercomputer LEONARDO, hosted by CINECA (Italy) and the LEONARDO consortium through an EuroHPC AI Access call.
Additionally, this work was granted access to the HPC resources of IDRIS under the allocation 2025-A0171015718 made by GENCI.
We thank Denis Habip Gatenyo and Jordan Mukadi for their help on the human evaluation.

\clearpage
\bibliography{main, raidium}

\clearpage

\begin{appendices}

\clearpage
\section{Extended Data}

\begin{figure*}[h]
    \centering
    \includegraphics[width=0.95\linewidth]{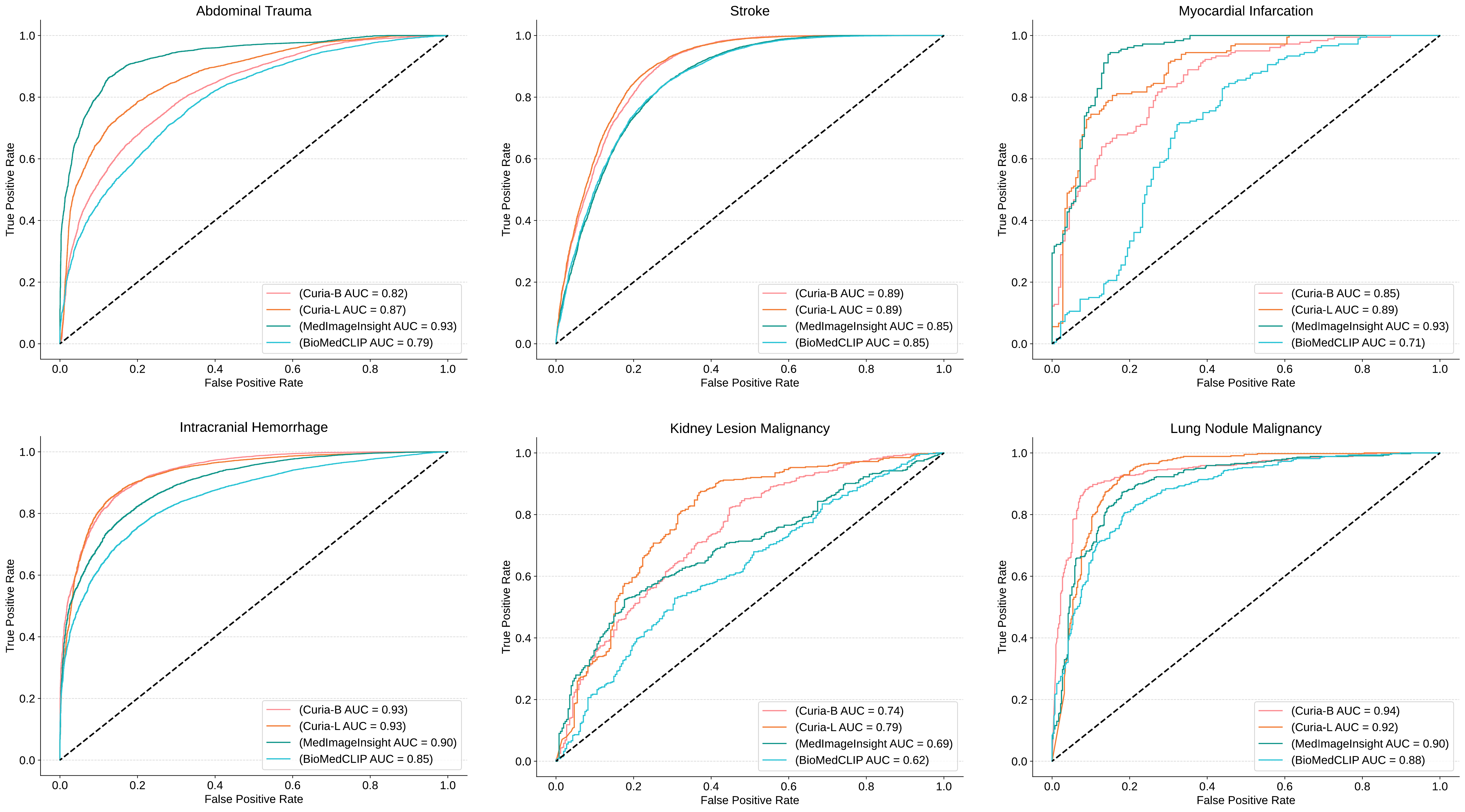}
    \caption{Receiver Operating Characteristic (ROC) curves for all binary classification tasks in our benchmark. The curves were computed on 5 runs for each models, and were aggregated using the pooling method~\cite{swets2012evaluation, hogan2023averaging}. All predictions from the 5 runs were concatenated into a single ensemble of predictions, that was used to plot the ROC curve.}
    \label{fig:roc}
\end{figure*}

\begin{figure*}[h]
    \centering
    \includegraphics[width=1.0\linewidth]{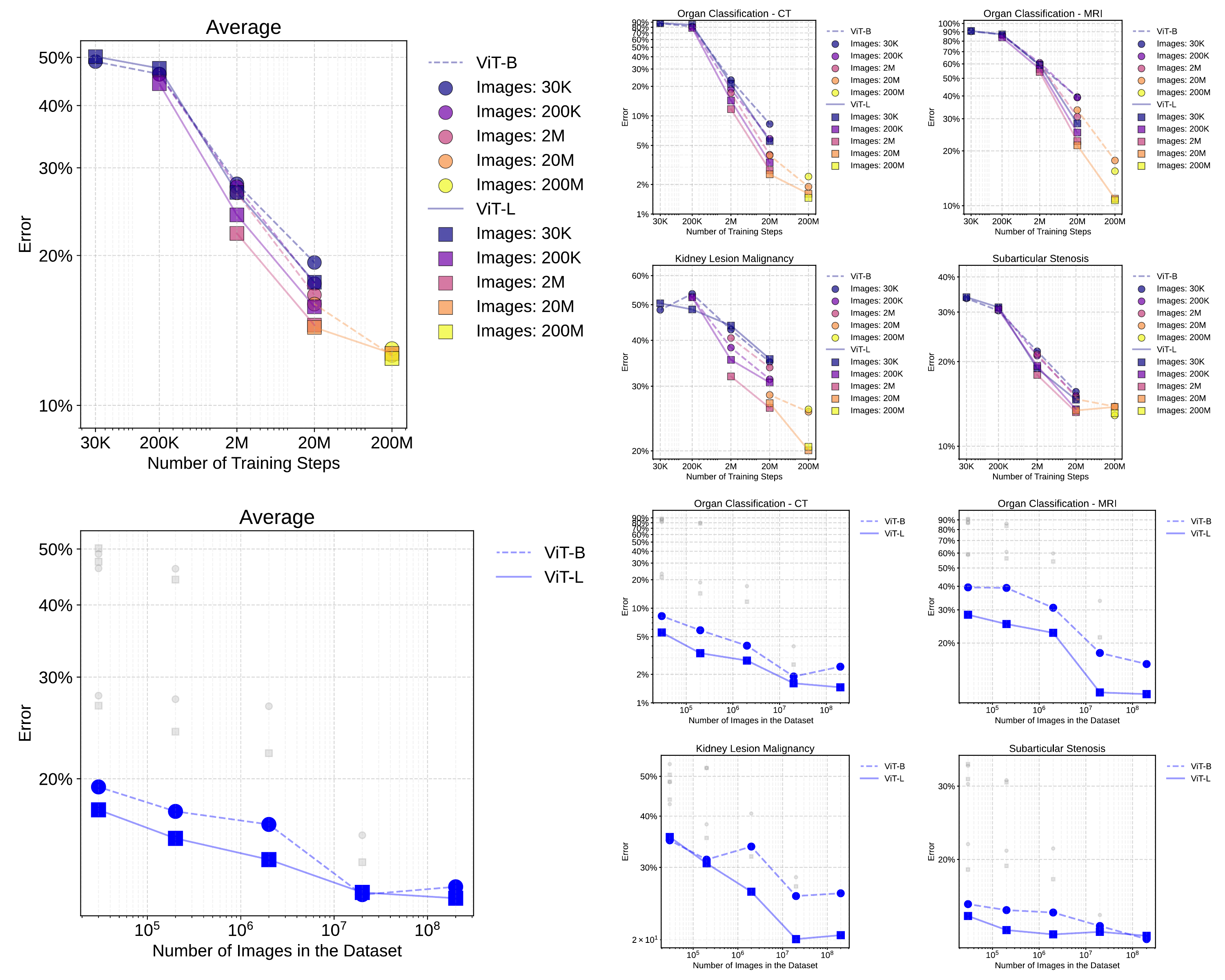}
    \caption{Scaling curves for two architectures: ViT-B and ViT-L. We train the two models with various dataset sizes (30K, 200K, 2M, 20M, 200M), for different number of training steps, and show all results in the top 5 plots.
    The average is computed on all our 19 benchmarks. On the bottom, we plot, for each dataset size, the best performing model.}
    \label{fig:scaling}
\end{figure*}

We display in Fig.~\ref{fig:scaling} results on our benchmark for multiple model sizes, dataset sizes and number of training steps. We used dataset sizes of 30K, 200K, 2M, 20M, and 200M.
The number of training steps followed the dataset sizes: for given global batch size b, we trained the model 30K/b  200K/b, 2M/b, 20M/b, and 200M/b steps: this means a model trained for 30K/b steps would have seen 30K images during its training.
We skipped the setups where the number of steps would result in not seeing the full dataset (e.g. 2M dataset with 30K/b steps).

We observe that increasing one of those three parameter, fixing the two others, leads to an increase in performance.
The number of training steps in particularly crucial: even on a small dataset, of 40K images, a ViT-B can reach an error rate of under 20\%

\clearpage
\section{Model Parameters}
\subsection{Pre-training hyperparameters}

The Table~\ref{tab:hyperparameters} presents the main hyperparameters used in the pre-training.

\begin{table*}[h!]
\centering
\caption{\textbf{Main hyperparameters used in the pre-training for \curia-B and \curia-L.}}\label{tab:hyperparameters}
\begin{tabular}{llll}
\toprule
Models                  &   & \curia-B  & \curia-L\\
\midrule
 \textbf{Optimization}  & Warmup iterations & 25,000 & 25,000 \\
                        & Optimizer & AdamW & AdamW \\
                        & Lr Scheduler & Cosine & Cosine \\
    \textbf{Parameters} & Weight decay start value & 0.04 & 0.04 \\
                        & Weight decay end value & 0.2 & 0.2 \\
                        & Total batch size & 512 & 256 \\
                        & Number of iterations & 475,000 & 475,000 \\
    \textbf{Model}      & Patch size & 16 & 16 \\
                        & Resolution & 512 & 512 \\
    \textbf{Parameters} & Register tokens & 0 & 0 \\
                        & Embedding dimension & 768 & 1024 \\
                        & Layers & 12 & 24 \\
                        & Heads & 12 & 16 \\
                        & MLP ratio & 4.0 & 4.0 \\
                        & MLP activation & SwiGLU fused & SwiGLU fused \\
    \textbf{Projection} & Heads prototypes & 131072 & 131072 \\
    \textbf{Heads}      & DINO head bottleneck dim & 384 & 384 \\
                        & iBOT head bottleneck dim & 256 & 256 \\
\textbf{Augmentation}   & Global crop scale & [0.32, 1.0] & [0.32, 1.0] \\
                        & Local crop scale & [0.05, 0.32] & [0.05, 0.32] \\
                        & Global crop number & 2 & 2 \\
                        & Local crop number & 8 & 8 \\
                        & Global crop size & 512 & 512 \\
                        & Local crop size & 224 & 224 \\
    \textbf{Hardware}   & GPUs & 16xA100 80 GB & 16xA100 80 GB\\
                        & Precision & FP16 & FP16 \\
\bottomrule
\end{tabular}
\end{table*}

\subsection{Head setups}

We provide in Table~\ref{tab:head} the different configurations used for each downstream tasks (see Fig. ~\ref{fig:classification-method}).

\begin{table*}[h!]
\centering
\caption{\textbf{Head setup for all tasks for \curia models.} We report whether features were aggregated at the image level or within segmentation masks (see Fig. \ref{fig:classification-method}). Additionally, we specify whether CLS tokens or patch tokens were used.}\label{tab:head}
\begin{tabular}{llllcc}
\toprule
Category & Benchmark & Level & Type & CLS Tokens & Patch Tokens\\
\midrule
\textbf{Anatomy} & CT Organ Recognition & mask & linear &  & \checkmark  \\
& MRI Organ Recognition & mask & linear &  & \checkmark  \\
& Neuroimaging Brain Estimation & image & linear & \checkmark &   \\
\textbf{Oncology} & Lung Nodule Malignancy & mask & attention &  & \checkmark  \\
& Kidney Lesion Malignancy & mask & linear &  & \checkmark  \\
& Tumor Localisation & mask & linear &  & \checkmark  \\
\textbf{Musculoskeletal} & Renal Malignancy Survival & mask & linear &  & \checkmark  \\
& Foraminal Narrowing & mask & linear &  & \checkmark  \\
& Spinal Cord Stenosis & mask & linear &   & \checkmark  \\
& Subarticular Stenosis & mask & linear &   & \checkmark  \\
& ACL Tear & mask & attention &  & \checkmark  \\
\textbf{Emergency} & Myocardial Infarction & mask & linear &  & \checkmark  \\
& Abdominal Trauma & image & linear & \checkmark &  \\
& Intracranial Hemorrhage & image & linear & \checkmark & \checkmark  \\
& Stroke & image & attention &  & \checkmark  \\
\textbf{Degenerative} & Alzheimer's Disease & image & attention & \checkmark & \checkmark  \\
\textbf{Infectious} & Pulmonary Infections & image & linear & \checkmark &   \\
\bottomrule
\end{tabular}
\end{table*}

\section{Prompted Segmentation}

Table~\ref{tab:prompted_seg} presents the detailed results of the prompted segmentation, reported per organ.

\begin{table*}[h]
\centering
\caption{\textbf{Results on the AMOS Dataset for Prompted Organ Segmentation.} For each model, we report the mean Dice Similarity Coefficient (DSC) computed per organ. Both bounding box (Bbox) and point-based prompts are considered in the evaluation.}
\label{tab:prompted_seg}
\begin{tabular}{lcccccccc}
\toprule
\multirow{3}{*}{Organ} & \multicolumn{2}{c}{SAM} & \multicolumn{2}{c}{RadSAM} & \multicolumn{2}{c}{Curia-B} & \multicolumn{2}{c}{Curia-L} \\
\cmidrule(lr){2-3}
\cmidrule(lr){4-5}
\cmidrule(lr){6-7}
\cmidrule(lr){8-9}
 & Bbox $\uparrow$ & Point $\uparrow$ & Bbox $\uparrow$ & Point $\uparrow$ & Bbox $\uparrow$ & Point $\uparrow$ & Bbox $\uparrow$ & Point $\uparrow$\\
\midrule
Aorta & 78.31 & 65.68 & 95.82 & 94.97 & 95.96 & 95.23 & 96.11 & 95.68 \\
Bladder & 76.08 & 30.98 & 93.29 & 90.50 & 92.85 & 90.17 & 93.47 & 91.32 \\
Duodenum & 57.43 & 18.07 & 85.97 & 70.04 & 85.82 & 74.78 & 86.39 & 76.99 \\
Esophagus & 59.44 & 6.60 & 88.28 & 83.47 & 87.95 & 85.67 & 88.49 & 83.67 \\
Gall Bladder & 76.50 & 26.97 & 91.51 & 80.22 & 90.89 & 83.45 & 91.08 & 77.84 \\
Left Adrenal Gland & 49.03 & 6.29 & 81.49 & 73.61 & 81.08 & 78.08 & 81.64 & 76.31 \\
Left Kidney & 87.91 & 80.70 & 96.51 & 95.84 & 96.56 & 96.51 & 96.75 & 96.66 \\
Liver & 81.59 & 50.24 & 97.03 & 94.50 & 97.56 & 95.96 & 97.77 & 96.64 \\
Pancreas & 63.12 & 26.12 & 85.92 & 75.03 & 87.77 & 82.45 & 87.97 & 84.49 \\
Postcava & 71.56 & 8.74 & 91.47 & 86.38 & 91.45 & 87.87 & 91.76 & 86.01 \\
Prostate Uterus & 70.73 & 17.65 & 91.56 & 84.31 & 91.10 & 88.13 & 91.71 & 89.17 \\
Right Adrenal Gland & 33.47 & 1.48 & 79.25 & 72.51 & 80.27 & 77.75 & 80.20 & 72.47 \\
Right Kidney & 85.40 & 71.01 & 96.49 & 95.52 & 96.44 & 95.75 & 96.63 & 95.91 \\
Spleen & 86.52 & 54.54 & 97.12 & 95.74 & 97.01 & 95.53 & 97.38 & 96.56 \\
Stomach & 76.11 & 30.18 & 94.64 & 86.08 & 94.31 & 90.50 & 95.10 & 92.21 \\
\midrule
All & 70.16 & 33.04 & 91.08 & 85.27 & 91.13 & \textbf{87.87} & \textbf{91.49} & 87.49 \\
\bottomrule
\end{tabular}
\end{table*}

\section{Detailed Results}\label{secA1}

\subsection{Registration}
This section showcases the results on the \textbf{image registration} benchmark. More specifically, Table~\ref{tab:registration-learn2reg-mrct} shows the results on Learn2Reg Abdomen, Table~\ref{tab:registration-learn2reg-oasis} the results on Learn2Reg Brain, and Table~\ref{tab:registration-gan-mrct} the results on XCAT.

\begin{table*}[h]
    \centering
    \caption{\textbf{Learn2Reg Abdomen MRI/CT Registration Results.} For each model, the metrics reported are, in order, the mean Dice Similiary Coefficient (DSC) score (in \%), the DSC scores on  liver, spleen, right kidney, and left kidney (in\%), and the standard deviation of the log-Jacobian determinant.}
    \label{tab:registration-learn2reg-mrct}
    \begin{tabular}{lrrrrrr}
    \toprule
       Model & Mean $\uparrow$ & Liver $\uparrow$ & Spleen $\uparrow$ & R Kidney $\uparrow$ & L Kidney $\uparrow$ & stdLogJ $\downarrow$ \\
    \midrule
        \curia-B        & \textbf{85.1} & \textbf{87.96} & 84.22 & \textbf{82.76} & \textbf{85.46} & \textbf{0.1039} \\
        \curia-L        & 83.84          & 86.11          & \textbf{84.27} & 81.55          & 83.41          & 0.3317 \\
        MedImageInsight & 77.83          & 75.54          & 74.38          & 80.24          & 81.16          & 0.3439 \\
        BiomedCLIP      & 74.52          & 83.99          & 74.07          & 71.66          & 68.36          & 0.1317 \\
        DINOv2 Large     & 79.83          & 84.03          & 74.55          & 80.51          & 80.26          & 0.1173 \\
    \bottomrule
    \end{tabular}
\end{table*}

\begin{table*}[h]
    \centering
    \caption{\textbf{Learn2Reg Brain Registration Results.} For each model, the metrics reported are, the mean Dice Similiary Coefficient (DSC) score (in \%), and the standard deviation of the log-Jacobian determinant.}
    \label{tab:registration-learn2reg-oasis}
    \begin{tabular}{lrrrrrr}
        \toprule
        Model & Mean $\uparrow$  & stdLogJ $\downarrow$	\\
        \midrule
        \curia-B  & 77.68 & \textbf{0.0519} \\
        \curia-L & \textbf{77.96} & 0.0938 \\
        MedImageInsight  & 75.91 & 0.0843 \\
        BiomedCLIP   & 76.29 & 0.1333 \\
        DINOv2 Large & 68.06 & 0.1572 \\
        \bottomrule
    \end{tabular}
\end{table*}

\begin{table*}[h]
    \centering
    \caption{\textbf{XCAT - Multimodal Abdominal Medical Image Registration Results.} For each model, the metrics reported are, in order, the mean Dice Similiary Coefficient (DSC) score (in \%), the DSC scores on  liver, spleen, right kidney, and left kidney (in\%), and the standard deviation of the log-Jacobian determinant.}
    \label{tab:registration-gan-mrct}
    \begin{tabular}{lrrrrrr}
    \toprule
    Model & Mean $\uparrow$ & Liver $\uparrow$ & Spleen $\uparrow$ & R Kidney $\uparrow$ & L Kidney $\uparrow$ & stdLogJ $\downarrow$ \\
    \midrule
    \curia-B (CT $\xrightarrow{}$ CT)         & 81.30 & 94.26 & 87.18 & \textbf{58.51} & \textbf{85.27} & 0.0369 \\
    \curia-L (CT $\xrightarrow{}$ CT)         & 80.12 & 93.37 & 87.30 & 56.46 & 83.35 & 0.0817 \\
    MedImageInsight (CT $\xrightarrow{}$ CT)  & 76.25 & 91.92 & 80.63 & 52.12 & 80.33 & \textbf{0.0292} \\
    BiomedCLIP (CT $\xrightarrow{}$ CT)       & \textbf{81.74} & \textbf{94.65} & \textbf{90.22} & 57.52 & 84.55 & 0.0603 \\
    DINOv2 Large (CT $\xrightarrow{}$ CT)       & 79.60 & 93.22 & 86.14 & 56.88 & 82.17 & 0.0615 \\
    \midrule
    \curia-B (MR $\xrightarrow{}$ MR)         & \textbf{86.10} & \textbf{94.47} & 84.66 & \textbf{82.91} & \textbf{82.38} & 0.0371 \\
    \curia-L (MR $\xrightarrow{}$ MR)         & 84.25 & 92.81 & 82.12 & 81.53 & 80.52 & 0.0722 \\
    MedImageInsight (MR $\xrightarrow{}$ MR)  & 76.55 & 88.02 & 70.51 & 73.25 & 74.40 & \textbf{0.0281} \\
    BiomedCLIP (MR $\xrightarrow{}$ MR)       & 82.96 & 94.25 & 80.12 & 79.52 & 77.96 & 0.0646 \\
    DINOv2 Large (MR $\xrightarrow{}$ MR)       & 85.81 & 93.99 & \textbf{86.13} & 82.89 & 80.24 & 0.0619 \\
    \midrule
    \curia-B (CT $\xrightarrow{}$ MR)         & 64.03 & \textbf{86.12} & 70.09 & 43.85 & 56.06 & 0.0633 \\
    \curia-L (CT $\xrightarrow{}$ MR)         & \textbf{65.25} & 85.34 & \textbf{71.92} & 44.41 & \textbf{59.33} & 0.1070 \\
    MedImageInsight (CT $\xrightarrow{}$ MR)  & 56.99 & 78.44 & 65.31 & 35.85 & 48.37 & \textbf{0.0468} \\
    BiomedCLIP (CT $\xrightarrow{}$ MR)       & 52.32 & 81.69 & 60.02 & 30.53 & 37.05 & 0.2233 \\
    DINOv2 Large (CT $\xrightarrow{}$ MR)       & 64.71 & 86.22 & 71.90  & \textbf{44.77} & 55.96 & 0.0843 \\
    \bottomrule
    \end{tabular}
\end{table*}

\subsection{Cross-Modality Generalization}
The results in Table~\ref{tab:AMOS-OoD} showcase the generalization capability of FMs in a cross-modality context. Models were fine-tuned on the organ recognition task using either CT or MRI data, and evaluated on the other modality—i.e., MRI or CT, respectively.

\begin{table*}[h]
    \centering
    \caption{\textbf{Cross-modality generalization results of the Organ Recognition task on CT and MRI.} The metrics reported are the balanced accuracies (in \%) on the 41 common classes between CT and MRI data in the benchmark.}
    \label{tab:AMOS-OoD}
    \begin{tabular}{lcccc}
    \toprule
    \multirow{3}{*}{Model} & \multicolumn{2}{c}{CT $\rightarrow$ MRI} & \multicolumn{2}{c}{MRI $\rightarrow$ CT} \\
    \cmidrule(l){2-3} \cmidrule(l){4-5}
     & \shortstack{In-Distribution\\ CT} & \shortstack{Out-of-Distribution\\ MRI} & \shortstack{In-Distribution\\ MRI} & \shortstack{Out-of-Distribution\\ CT} \\
    \midrule
    Curia-L & 97.44 & 88.27 & 95.79 & 96.40\\
    BiomedCLIP & 85.62 & 42.53 & 72.30 & 54.99\\
    MedImageInsight & 88.59 & 53.08 & 70.66 & 63.52\\
    ViT-L & 84.29 & 12.53 & 31.18 & 18.04\\
     \bottomrule
    \end{tabular}
\end{table*}

\section{Detailed Scores by Benchmark}

\subsection{Bootstrapping Scores}
\label{app:scores}
In this section, we display the detailed scores for each benchmark obtained through non-parametric bootstrapping. Specifically, we report the average value of the main metric along with the corresponding 95\% confidence intervals.
\begin{table*}[h!]
\centering
\caption{\textbf{Comparison of Models for Benchmark: CT Organ Recognition.} Each model’s performance was evaluated over 5 training runs using non-parametric bootstrapping with 1,000 resamples. The table reports the average accuracy score along with the corresponding 95\% confidence intervals computed across the bootstrapped distribution (all metrics in \%).}
\begin{tabular}{lccc}
\toprule
Model & Accuracy Score $\uparrow$ & Lower 95\% CI
 & Upper 95\% CI
 \\
\midrule
Curia-B & 98.10 & 97.58 & 98.55 \\
Curia-L & 98.40 & 97.93 & 98.83 \\
MedImageInsight & 88.19 & 87.00 & 89.33 \\
BiomedCLIP & 84.95 & 83.75 & 86.07 \\
\bottomrule
\end{tabular}
\end{table*}

\begin{table*}[h!] \centering
\caption{\textbf{Comparison of Models for Benchmark: MRI Organ Recognition.} Each model’s performance was evaluated over 5 training runs using non-parametric bootstrapping with 1,000 resamples. The table reports the average accuracy score along with the corresponding 95\% confidence intervals computed across the bootstrapped distribution.}
\begin{tabular}{lccc}
\toprule
Model & Accuracy Score $\uparrow$ & Lower 95\% CI
 & Upper 95\% CI \\
\midrule
Curia-B & 82.27 & 80.22 & 84.16 \\
Curia-L & 89.11 & 87.59 & 90.69 \\
MedImageInsight & 63.18 & 60.62 & 65.66 \\
BiomedCLIP & 63.22 & 60.78 & 65.56 \\
\bottomrule
\end{tabular}
\end{table*}

\begin{table*}[h!] \centering
\caption{\textbf{Comparison of Models for Benchmark: Neuroimaging Brain Estimation.} Each model’s performance was evaluated over 5 training runs using non-parametric bootstrapping with 1,000 resamples. The table reports the average accuracy score along with the corresponding 95\% confidence intervals computed across the bootstrapped distribution.}
\begin{tabular}{lccc}
\toprule
Model & $\text{r}^2$ Score $\uparrow$ & Lower 95\% CI
 & Upper 95\% CI \\
\midrule
Curia-B & 75.80 & 73.35 & 78.17 \\
Curia-L & 75.54 & 72.94 & 77.66 \\
MedImageInsight & 72.46 & 69.69 & 75.17 \\
BiomedCLIP & 69.41 & 66.35 & 71.85 \\
\bottomrule
\end{tabular}
\end{table*}

\begin{table*}[h!] \centering
\caption{\textbf{Comparison of Models for Benchmark: Lung Nodule Malignancy.} Each model’s performance was evaluated over 5 training runs using non-parametric bootstrapping with 1,000 resamples. The table reports the average accuracy score along with the corresponding 95\% confidence intervals computed across the bootstrapped distribution. * indicates results reported from the original publication. While all other models only train a linear head, Harvard OncoFM finetuned also updates the pretrained encoder weights on the downstream task.}
\begin{tabular}{lccc}
\toprule
Model & AUC Score $\uparrow$ & Lower 95\% CI
 & Upper 95\% CI \\
\midrule
Curia-B & 94.98 & 92.05 & 97.41 \\
Curia-L & 92.45 & 87.96 & 96.26 \\
MedImageInsight & 92.18 & 88.19 & 95.70 \\
BiomedCLIP & 88.72 & 83.99 & 92.82 \\
Harvard OncoFM* & 88.23 & - & - \\
Harvard OncoFM finetuned* & 94.40 & - & - \\
\bottomrule
\end{tabular}
\end{table*}

\begin{table*}[h!] \centering
\caption{\textbf{Comparison of Models for Benchmark: Kidney Lesion Malignancy.} Each model’s performance was evaluated over 5 training runs using non-parametric bootstrapping with 1,000 resamples. The table reports the average accuracy score along with the corresponding 95\% confidence intervals computed across the bootstrapped distribution.}
\begin{tabular}{lccc}
\toprule
Model & AUC Score $\uparrow$ & Lower 95\% CI
 & Upper 95\% CI \\
\midrule
Curia-B & 74.41 & 66.08 & 82.10 \\
Curia-L & 80.29 & 72.16 & 87.83 \\
MedImageInsight & 67.81 & 59.62 & 76.20 \\
BiomedCLIP & 62.95 & 54.62 & 71.05 \\
\bottomrule
\end{tabular}
\end{table*}

\begin{table*}[h!] \centering
\caption{\textbf{Comparison of Models for Benchmark: Tumor Localisation.} Each model’s performance was evaluated over 5 training runs using non-parametric bootstrapping with 1,000 resamples. The table reports the average accuracy score along with the corresponding 95\% confidence intervals computed across the bootstrapped distribution.}
\begin{tabular}{lccc}
\toprule
Model & Accuracy Score $\uparrow$ & Lower 95\% CI
 & Upper 95\% CI \\
\midrule
Curia-B & 91.87 & 89.76 & 93.75 \\
Curia-L & 91.74 & 89.56 & 93.84 \\
MedImageInsight & 88.91 & 85.99 & 91.40 \\
BiomedCLIP & 86.39 & 83.68 & 88.95 \\
\bottomrule
\end{tabular}
\end{table*}

\begin{table*}[h!] \centering
\caption{\textbf{Comparison of Models for Benchmark: Renal Malignancy Survival.} Each model’s performance was evaluated over 5 training runs using non-parametric bootstrapping with 1,000 resamples. The table reports the average accuracy score along with the corresponding 95\% confidence intervals computed across the bootstrapped distribution.}
\begin{tabular}{lccc}
\toprule
Model & c-index Score $\uparrow$ & Lower 95\% CI
 & Upper 95\% CI \\
\midrule
Curia-B & 71.12 & 65.38 & 76.86 \\
Curia-L & 62.81 & 56.43 & 69.52 \\
MedImageInsight & 62.79 & 56.03 & 69.24 \\
BiomedCLIP & 63.94 & 57.02 & 70.15 \\
\bottomrule
\end{tabular}
\end{table*}

\begin{table*}[h!] \centering
\caption{\textbf{Comparison of Models for Benchmark: Foraminal
Narrowing.} Each model’s performance was evaluated over 5 training runs using non-parametric bootstrapping with 1,000 resamples. The table reports the average accuracy score along with the corresponding 95\% confidence intervals computed across the bootstrapped distribution.}
\begin{tabular}{lccc}
\toprule
Model & AUC Score $\uparrow$ & Lower 95\% CI
 & Upper 95\% CI \\
\midrule
Curia-B & 86.16 & 84.71 & 87.69 \\
Curia-L & 86.21 & 84.76 & 87.62 \\
MedImageInsight & 86.32 & 84.64 & 87.75 \\
BiomedCLIP & 84.45 & 82.93 & 86.13 \\
\bottomrule
\end{tabular}
\end{table*}

\begin{table*}[h!] \centering
\caption{\textbf{Comparison of Models for Benchmark: Spinal Canal Stenosis.} Each model’s performance was evaluated over 5 training runs using non-parametric bootstrapping with 1,000 resamples. The table reports the average accuracy score along with the corresponding 95\% confidence intervals computed across the bootstrapped distribution.}
\begin{tabular}{lccc}
\toprule
Model & AUC Score $\uparrow$ & Lower 95\% CI
 & Upper 95\% CI \\
\midrule
Curia-B & 94.65 & 93.22 & 95.94 \\
Curia-L & 93.73 & 92.38 & 95.00 \\
MedImageInsight & 92.98 & 91.31 & 94.58 \\
BiomedCLIP & 92.33 & 90.61 & 93.91 \\
\bottomrule
\end{tabular}
\end{table*}

\begin{table*}[h!] \centering
\caption{\textbf{Comparison of Models for Benchmark: Subarticular Stenosis.} Each model’s performance was evaluated over 5 training runs using non-parametric bootstrapping with 1,000 resamples. The table reports the average accuracy score along with the corresponding 95\% confidence intervals computed across the bootstrapped distribution.}
\begin{tabular}{lrrr}
\toprule
Model & AUC Score $\uparrow$ & Lower 95\% CI
 & Upper 95\% CI \\
\midrule
Curia-B & 87.81 & 89.09 & 86.57 \\
Curia-L & 87.46 & 86.14 & 88.72 \\
MedImageInsight & 85.61 & 84.26 & 87.00 \\
BiomedCLIP & 83.92 & 82.53 & 85.38 \\
\bottomrule
\end{tabular}
\end{table*}

\begin{table*}[h!] \centering
\caption{\textbf{Comparison of Models for Benchmark: ACL Tear.} Each model’s performance was evaluated over 5 training runs using non-parametric bootstrapping with 1,000 resamples. The table reports the average accuracy score along with the corresponding 95\% confidence intervals computed across the bootstrapped distribution.}
\begin{tabular}{lccc}
\toprule
Model & AUC Score $\uparrow$ & Lower 95\% CI
 & Upper 95\% CI \\
\midrule
Curia-B & 85.03 & 77.91 & 91.80 \\
Curia-L & 87.34 & 80.89 & 92.97 \\
MedImageInsight & 78.39 & 68.62 & 86.52 \\
BiomedCLIP & 81.97 & 75.29 & 88.32 \\
\bottomrule
\end{tabular}
\end{table*}

\begin{table*}[h!] \centering
\caption{\textbf{Comparison of Models for Benchmark: Myocardial Infarction.} Each model’s performance was evaluated over 5 training runs using non-parametric bootstrapping with 1,000 resamples. The table reports the average accuracy score along with the corresponding 95\% confidence intervals computed across the bootstrapped distribution.}
\begin{tabular}{lccc}
\toprule
Model & AUC Score $\uparrow$ & Lower 95\% CI
 & Upper 95\% CI \\
\midrule
Curia-B & 84.55 & 75.92 & 91.98 \\
Curia-L & 89.16 & 81.40 & 95.65 \\
MedImageInsight & 94.08 & 87.87 & 98.80 \\
BiomedCLIP & 71.39 & 58.85 & 82.44 \\
\bottomrule
\end{tabular}
\end{table*}

\begin{table*}[h!] \centering
\caption{\textbf{Comparison of Models for Benchmark: Abdominal Trauma.} Each model’s performance was evaluated over 5 training runs using non-parametric bootstrapping with 1,000 resamples. The table reports the average accuracy score along with the corresponding 95\% confidence intervals computed across the bootstrapped distribution.}
\begin{tabular}{lccc}
\toprule
Model & AUC Score $\uparrow$ & Lower 95\% CI
 & Upper 95\% CI \\
\midrule
Curia-B & 82.63 & 83.98 & 81.31 \\
Curia-L & 87.10 & 85.75 & 88.28 \\
MedImageInsight & 93.14 & 92.27 & 93.99 \\
BiomedCLIP & 79.14 & 77.58 & 80.60 \\
\bottomrule
\end{tabular}
\end{table*}

\begin{table*}[h!] \centering
\caption{\textbf{Comparison of Models for Benchmark: Intracranial Hemorrhage.} Each model’s performance was evaluated over 5 training runs using non-parametric bootstrapping with 1,000 resamples. The table reports the average accuracy score along with the corresponding 95\% confidence intervals computed across the bootstrapped distribution.}
\begin{tabular}{lccc}
\toprule
Model & AUC Score $\uparrow$ & Lower 95\% CI
 & Upper 95\% CI \\
\midrule
Curia-B & 93.69 & 92.77 & 94.56 \\
Curia-L & 93.54 & 92.69 & 94.40 \\
MedImageInsight & 90.11 & 88.94 & 91.19 \\
BiomedCLIP & 87.77 & 86.57 & 88.92 \\
\bottomrule
\end{tabular}
\end{table*}

\begin{table*}[h!] \centering
\caption{\textbf{Comparison of Models for Benchmark: Stroke.} Each model’s performance was evaluated over 5 training runs using non-parametric bootstrapping with 1,000 resamples. The table reports the average accuracy score along with the corresponding 95\% confidence intervals computed across the bootstrapped distribution.}
\begin{tabular}{lccc}
\toprule
Model & AUC Score $\uparrow$ & Lower 95\% CI
 & Upper 95\% CI \\
\midrule
Curia-B & 89.93 & 89.02 & 90.83 \\
Curia-L & 89.78 & 88.79 & 90.67 \\
MedImageInsight & 88.62 & 87.47 & 89.66 \\
BiomedCLIP & 85.72 & 84.52 & 86.93 \\
\bottomrule
\end{tabular}
\end{table*}

\begin{table*}[h!] \centering
\caption{\textbf{Comparison of Models for Benchmark: Alzheimer's Disease.} Each model’s performance was evaluated over 5 training runs using non-parametric bootstrapping with 1,000 resamples. The table reports the average accuracy score along with the corresponding 95\% confidence intervals computed across the bootstrapped distribution.}
\begin{tabular}{lccc}
\toprule
Model & AUC Score $\uparrow$ & Lower 95\% CI
 & Upper 95\% CI \\
\midrule
Curia-B & 87.83 & 77.70 & 95.33 \\
Curia-L & 84.90 & 74.51 & 93.78 \\
MedImageInsight & 87.66 & 75.78 & 96.38 \\
BiomedCLIP & 88.19 & 77.36 & 96.45 \\
\bottomrule
\end{tabular}
\end{table*}

\begin{table*}[h!] \centering
\caption{\textbf{Comparison of Models for Benchmark: Pulmonary Infections.} Each model’s performance was evaluated over 5 training runs using non-parametric bootstrapping with 1,000 resamples. The table reports the average accuracy score along with the corresponding 95\% confidence intervals computed across the bootstrapped distribution.}
\begin{tabular}{lccc}
\toprule
Model & Balanced Accuracy Score $\uparrow$ & Lower 95\% CI
 & Upper 95\% CI \\
\midrule
Curia-B & 91.49 & 90.54 & 92.43 \\
Curia-L & 93.40 & 92.61 & 94.18 \\
MedImageInsight & 89.97 & 88.83 & 91.02 \\
BiomedCLIP & 89.25 & 88.22 & 90.26 \\
\bottomrule
\end{tabular}
\end{table*}

\subsection{Statistical Significance}
\label{app:pvalues}
In this section, we present the results of statistical significance tests performed using a paired bootstrap approach across five training runs. Specifically, for each pair of models and for each benchmark, we report the computed p-value as well as the 95\% confidence interval (in \%) of the main metric, derived from the paired bootstrap distribution.
\begin{table*}
\centering
\caption{\textbf{Statistical Comparison — Curia-B vs MedImageInsight.} For each benchmark and category, we report the p-value from a paired bootstrap test (1,000 resamples across 5 training runs) assessing whether the performance difference between the two models is statistically significant. The table also includes the 95\% confidence interval (in \%) of the main metric from the paired bootstrap distribution.}
\begin{tabular}{llccc}
\toprule
Category & Benchmark & p-value & Lower 95\% CI & Upper 95\% CI \\
\midrule
\textbf{Anatomy} & CT Organ Recognition & $<$ 0.001 & 8.84 & 10.94 \\
& MRI Organ Recognition & $<$ 0.001 & 16.74 & 21.48 \\
& Neuroimaging Brain Estimation & 0.006 & 1.20 & 5.80 \\
\textbf{Oncology} & Lung Nodule Malignancy & 0.126 & -0.01 & 0.06 \\
& Kidney Lesion Malignancy & 0.177 & -0.02 & 0.16 \\
& Tumor Localisation & 0.027 & 0.49 & 5.80 \\
& Renal Malignancy Survival & 0.003 & 2.96 & 13.64 \\
\textbf{Musculoskeletal} & Foraminal Narrowing & 0.812 & -0.01 & 0.01 \\
& Spinal Cord Stenosis & 0.011 & 0.00 & 0.03 \\
& Subarticular Stenosis & $<$ 0.001 & 1.41 & 3.02 \\
& ACL Tear & 0.093 & -0.79 & 15.99 \\
\textbf{Emergency} & Myocardial Infarction & 0.017 & -0.17 & -0.03 \\
& Abdominal Trauma & $<$ 0.001 & -0.12 & -0.09 \\
& Intracranial Hemorrhage & $<$ 0.001 & 0.03 & 0.04 \\
& Stroke & 0.001 & 0.01 & 0.02 \\
\textbf{Degenerative} & Alzheimer's Disease & 0.936 & -6.22 & 6.60 \\
\textbf{Infectious} & Pulmonary Infections & 0.007 & 0.40 & 2.61 \\
\bottomrule
\end{tabular}
\end{table*}

\begin{table*}
\centering
\caption{\textbf{Statistical Comparison — Curia-L vs MedImageInsight.} For each benchmark and category, we report the p-value from a paired bootstrap test (1,000 resamples across 5 training runs) assessing whether the performance difference between the two models is statistically significant. The table also includes the 95\% confidence interval (in \%) of the main metric from the paired bootstrap distribution.}
\begin{tabular}{llccc}
\toprule
Category & Benchmark & p-value & Lower 95\% CI & Upper 95\% CI \\
\midrule
\textbf{Anatomy} & CT Organ Recognition & $<$ 0.001 & 9.23 & 11.26 \\
& MRI Organ Recognition & $<$ 0.001 & 23.49 & 28.48 \\
& Neuroimaging Brain Estimation & 0.004 & 0.81 & 5.33 \\
\textbf{Oncology} & Lung Nodule Malignancy & 0.876 & -0.04 & 0.05 \\
& Kidney Lesion Malignancy & 0.025 & 0.02 & 0.22 \\
& Tumor Localisation & 0.041 & 0.17 & 5.59 \\
& Renal Malignancy Survival & 0.99 & -4.47 & 4.52 \\
\textbf{Musculoskeletal} & Foraminal Narrowing & 0.87 & -0.01 & 0.01 \\
& Spinal Cord Stenosis & 0.243 & -0.00 & 0.02 \\
& Subarticular Stenosis & $<$ 0.001 & 0.01 & 0.03 \\
& ACL Tear & 0.013 & 2.04 & 15.85 \\
\textbf{Emergency} & Myocardial Infarction & 0.104 & -0.12 & 0.01 \\
& Abdominal Trauma & $<$ 0.001 & -0.07 & -0.05 \\
& Intracranial Hemorrhage & $<$ 0.001 & 0.03 & 0.04 \\
& Stroke & 0.001 & 0.01 & 0.02 \\
\textbf{Degenerative} & Alzheimer's Disease & 0.325 & -7.88 & 2.63 \\
\textbf{Infectious} & Pulmonary Infections & $<$ 0.001 & 3.141 & 5.257 \\
\bottomrule
\end{tabular}
\end{table*}

\begin{table*}
\centering
\caption{\textbf{Statistical Comparison — Curia-B vs BiomedCLIP.} For each benchmark and category, we report the p-value from a paired bootstrap test (1,000 resamples across 5 training runs) assessing whether the performance difference between the two models is statistically significant. The table also includes the 95\% confidence interval (in \%) of the main metric from the paired bootstrap distribution.}
\begin{tabular}{llccc}
\toprule
Category & Benchmark & p-value & Lower 95\% CI & Upper 95\% CI \\
\midrule
\textbf{Anatomy} & CT Organ Recognition & $<$ 0.001 & 11.98 & 14.20 \\
& MRI Organ Recognition & $<$ 0.001 & 16.90 & 21.19 \\
& Neuroimaging Brain Estimation & $<$ 0.001 & 0.05 & 0.07 \\
\textbf{Oncology} & Lung Nodule Malignancy & 0.482 & -0.03 & 0.07 \\
& Kidney Lesion Malignancy & $<$ 0.001 & 3.81 & 9.11 \\
& Tumor Localisation & $<$ 0.001 & 1.21 & 3.34 \\
& Renal Malignancy Survival & 0.035 & 0.51 & 14.08 \\
\textbf{Musculoskeletal} & Foraminal Narrowing & 0.002 & 0.03 & 0.11 \\
& Spinal Cord Stenosis & 0.907 & -7.97 & 7.66 \\
& Subarticular Stenosis & $<$ 0.001 & 0.01 & 0.03 \\
& ACL Tear & 0.051 & 0.01 & 0.25 \\
\textbf{Emergency} & Myocardial Infarction & $<$ 0.001 & 2.83 & 8.06 \\
& Abdominal Trauma & $<$ 0.001 & 0.02 & 0.05 \\
& Intracranial Hemorrhage & 0.053 & -0.00 & 0.26 \\
& Stroke & $<$ 0.001 & 0.03 & 0.05 \\
\textbf{Degenerative} & Alzheimer's Disease & 0.003 & 0.53 & 2.87 \\
\textbf{Infectious} & Pulmonary Infections & $<$ 0.001 & 2.86 & 4.94 \\
\bottomrule
\end{tabular}
\end{table*}

\begin{table*}
\centering
\caption{\textbf{Statistical Comparison — Curia-L vs BiomedCLIP.} For each benchmark and category, we report the p-value from a paired bootstrap test (1,000 resamples across 5 training runs) assessing whether the performance difference between the two models is statistically significant. The table also includes the 95\% confidence interval (in \%) of the main metric from the paired bootstrap distribution.}
\begin{tabular}{llccc}
\toprule
Category & Benchmark & p-value & Lower 95\% CI & Upper 95\% CI \\
\midrule
\textbf{Anatomy} & CT Organ Recognition & $<$ 0.001 & 12.37 & 14.57 \\
& MRI Organ Recognition & $<$ 0.001 & 23.62 & 28.07 \\
& Neuroimaging Brain Estimation & $<$ 0.001 & 0.05 & 0.07 \\
\textbf{Oncology} & Lung Nodule Malignancy & 0.993 & -0.06 & 0.06 \\
& Kidney Lesion Malignancy & $<$ 0.001 & 3.48 & 8.75 \\
& Tumor Localisation & $<$ 0.001 & 3.14 & 5.26 \\
& Renal Malignancy Survival & 0.79 & -7.87 & 6.36 \\
\textbf{Musculoskeletal} & Foraminal Narrowing & 0.07 & -0.00 & 0.08 \\
& Spinal Cord Stenosis & 0.344 & -10.10 & 3.56 \\
& Subarticular Stenosis & 0.02 & 0.00 & 0.02 \\
& ACL Tear & 0.004 & 0.05 & 0.28 \\
\textbf{Emergency} & Myocardial Infarction & $<$ 0.001 & 2.68 & 7.89 \\
& Abdominal Trauma & $<$ 0.001 & 0.06 & 0.10 \\
& Intracranial Hemorrhage & 0.015 & 0.04 & 0.31 \\
& Stroke & $<$ 0.001 & 0.03 & 0.05 \\
\textbf{Degenerative} & Alzheimer's Disease & 0.005 & 0.50 & 2.94 \\
\textbf{Infectious} & Pulmonary Infections & $<$ 0.001 & 0.02 & 0.05 \\
\bottomrule
\end{tabular}
\end{table*}

\end{appendices}

\end{document}